\documentclass[10pt,twocolumn,letterpaper]{article}
\usepackage{cvpr}
\usepackage { makecell }
\usepackage{multirow}
\usepackage{amsmath,amsfonts}
\usepackage{algorithmic}
\usepackage{algorithm}
\usepackage{array}
\usepackage{xurl}
\usepackage{dsfont}
\usepackage{ulem}
\usepackage{amsfonts}
\usepackage{mathrsfs}
\usepackage{multirow}
\usepackage{colortbl}

\definecolor{cvprblue}{rgb}{0.21,0.49,0.74}
\usepackage[pagebackref,breaklinks,colorlinks,citecolor=cvprblue,urlcolor=cvprblue]{hyperref}

\def\paperID{XXXX}
\def\confName{CVPR}
\def\confYear{2026}

\usepackage[table]{xcolor}
\definecolor{rowlight}{RGB}{240,240,240}
\newcommand{\gc}[1]{\multicolumn{1}{>{\columncolor{rowlight}}c}{#1}}

\usepackage{array}
\usepackage[table]{xcolor}
\usepackage{makecell}
\usepackage{marvosym}
\definecolor{postgreen}{RGB}{214,236,214}
\definecolor{noliyellow}{RGB}{255,246,200}
\definecolor{headblue}{RGB}{213,232,252}

\usepackage{amsmath,amssymb,amsthm}
\usepackage{mathtools}
\newtheorem{proposition}{Proposition}

\newcommand{\egi}{\textit{e.g.}}
\newcommand{\iei}{\textit{i.e.}}

\newcommand{\settablefont}{\fontsize{7}{8}\selectfont}

\newcommand\clb[1]{{\color{black}{#1}}}
\newcommand\clc[1]{{\color{black}{#1}}}

\newcommand\clgray[1]{{\color{gray}{#1}}}

\usepackage{cuted}

\title{

The Midas Touch for Metric Depth
\vspace{-1.5em}
}

\author{
\begin{tabular*}{0.70\linewidth}{@{\extracolsep{\fill}}cccc}
Yu Ma\textsuperscript{1} &
Zizhan Guo\textsuperscript{1} &
Zuyi Xiong\textsuperscript{1} &
Haoran Zhang\textsuperscript{1} \\
Yi Feng\textsuperscript{1} &
Hongbo Zhao\textsuperscript{1} &
Hanli Wang\textsuperscript{1,2} &
Rui Fan\textsuperscript{1,2,3}\textsuperscript{\Letter}
\end{tabular*}\\[0.5em]
\vspace{-1.5em}
\and
\textsuperscript{1}\fontsize{11.3pt}{13pt}\selectfont College of Electronic and Information Engineering, Tongji University \\
\textsuperscript{2}\fontsize{11.3pt}{13pt}\selectfont Shanghai Research Institute for Intelligent Autonomous Systems, Tongji University \\
\textsuperscript{3}\fontsize{11.3pt}{13pt}\selectfont National Key Laboratory of Human-Machine Hybrid Augmented Intelligence, Xi'an Jiaotong University\\
}

\begin{document}

\maketitle

\begin{strip}
\vspace{-4.5em}
\centering
\includegraphics[width=\textwidth]{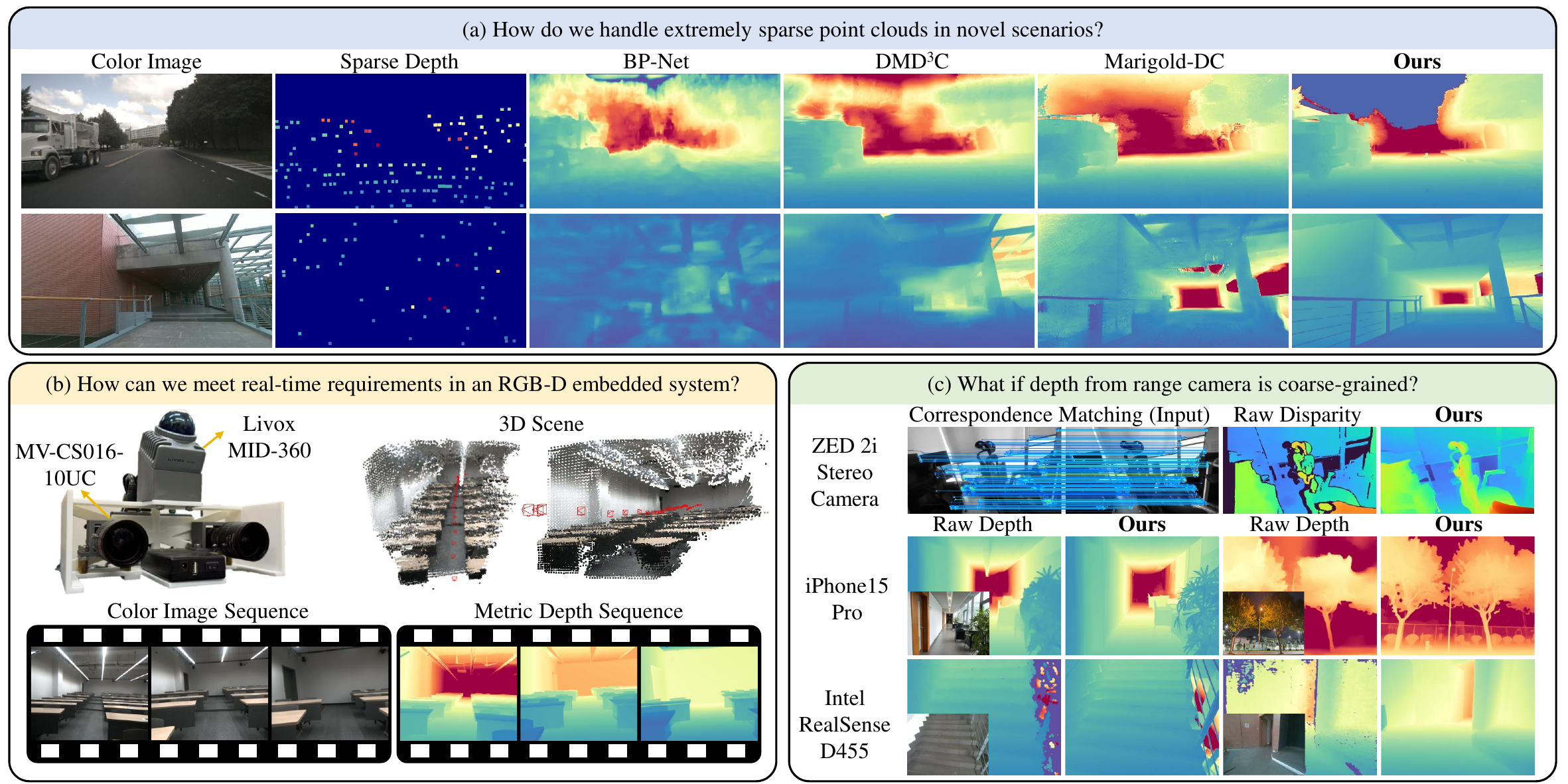}
\captionof{figure}{Application versatility of \clb{MTD} in metric depth \clb{perception}. (a) For novel scenes with extremely sparse point clouds, our method achieves precise depth completion and outperforms existing state-of-the-art methods. (b)
To eliminate offline LiDAR point cloud aggregation, our method achieves real-time, online predictions on handheld edge devices, thanks to its low inference time and high accuracy.
(c) For commonly used range cameras, our method can also serve as a plug-and-play module to enhance the quality of low-cost depth data.}
\label{fig:fig_1}
\end{strip}

\begingroup
\renewcommand\thefootnote{}
\footnotetext{\textsuperscript{\Letter}Corresponding author.}
\endgroup

\begin{abstract}
Recent advances have markedly improved the cross-scene generalization of relative depth estimation, yet its practical applicability remains limited by the absence of metric scale, local inconsistencies, and low computational efficiency. To address these issues, we present \emph{\textbf{M}idas \textbf{T}ouch for \textbf{D}epth} (MTD), a mathematically interpretable approach that converts relative depth into metric depth using only extremely sparse 3D data. To eliminate local scale inconsistencies, it applies a segment-wise recovery strategy via sparse graph optimization, followed by a pixel-wise refinement strategy using a discontinuity-aware geodesic cost. MTD exhibits strong generalization and achieves substantial accuracy improvements over previous depth completion and depth estimation methods. Moreover, its lightweight, plug-and-play design facilitates deployment and integration on diverse downstream 3D tasks. Project page is available at \url{https://mias.group/MTD}.

\end{abstract}

\section{Introduction}
\label{sect.intro}

\clc{

\textit{``In Greek mythology, King Midas’s ability to turn everything into gold is known as the `Midas touch'. Gold from a touch; meters from a hint. With extremely sparse 3D cues, relative depth crystallizes into metric measurement.''}

}

In recent years, monocular depth perception has emerged as a pivotal research focus in digital entertainment, computational photography, and 3D modeling~\cite{depthanythingv2,geobench}. Driven by the growing demand for strong generalization capabilities, including zero-shot performance, numerous efforts, such as the MiDaS series \cite{ranftl2020midas, birkl2023midasv3} and the DepthAnything series \cite{depthanything, depthanythingv2}, have focused on developing depth foundation models for relative depth estimation. Despite significant advancements in generalization ability, their practical applicability remains limited. This limitation arises not only from their large parameter counts and inference latency, but more fundamentally from the inherent scale ambiguity~\cite{yin2023metric3d}, which hinders accurate metric depth prediction.

To resolve this scale ambiguity problem, a straightforward approach in previous studies~\cite{ranftl2020midas, depthanything, marigold, lotus} leverages 3D point clouds to perform least-squares optimization for global scale recovery.
Nevertheless, this solution often suffers from limited precision, as local regions may exhibit scale inconsistencies: different instances or segments exhibit distinct scale ratios and shift biases utilized for recovering metric depth. Therefore, a single global rescaling cannot accommodate these variations, thereby degrading metric depth accuracy and downstream task performance.
Another line of research investigates incorporating captured 3D data into the metric depth estimation framework via networks~\cite{promptda, bpnet, dmd3c}. However, these approaches are typically trained in domain-specific environments, which limits their ability to generalize. Addressing this issue requires large-scale training datasets, which in turn significantly increase the cost and complexity of data collection.

Motivated by these limitations, we introduce the \emph{\textbf{\textit{Midas Touch for Depth}}} (MTD), a universal paradigm that leverages available 3D data to efficiently and accurately convert relative depth to metric depth. \clb{We avoid fine-tuning the depth foundation models, thereby preventing potential performance degradation.}
To maintain interpretability and improve efficiency, we develop a reliable nonparametric method with a clear mathematical foundation. Specifically, we adopt a parallel, segment-wise strategy that optimizes a sparse segment graph to correct local scale inconsistencies. To further compensate for pixel-level errors, we perform a pixel-wise refinement by reformulating depth propagation as a discontinuity-aware geodesic problem, thereby enabling an efficient dynamic-programming solution.

To validate the effectiveness and real-world applicability of our method, we conduct large-scale experiments across indoor and outdoor depth perception benchmarks. Our method achieves strong accuracy and generalization, outperforming state-of-the-art (SoTA) approaches, as shown in Fig.~\ref{fig:fig_1}. The results show that even under conditions of extremely sparse 3D data, our algorithm remains stable. To underscore the challenges posed by highly sparse, multi-source inputs (\egi, depth, disparity, or correspondence matches), we refer to them metaphorically as \emph{\textbf{\textit{3D seeds}}}, a term used throughout the paper. Moreover, our approach still recovers accurate metric depth, even when the relative depth predictions are only of moderate quality due to the reductions in model size. This advancement opens up new possibilities for improved computational efficiency. Our contributions are summarized as follows:

\begin{enumerate}

\item An efficient, effective, and universal paradigm for converting relative depth to metric depth, grounded in interpretable mathematical foundations;
\item A segment-wise scale recovery strategy via sparse graph optimization, complemented by a pixel-wise refinement strategy that uses a discontinuity-aware geodesic cost;
\item A SoTA, highly practical \clb{method} which can handle different types of extremely sparse 3D seeds and support various downstream 3D tasks.

\end{enumerate}

\begin{figure*}[!t]
	\centering
	\includegraphics[width=0.99\textwidth]{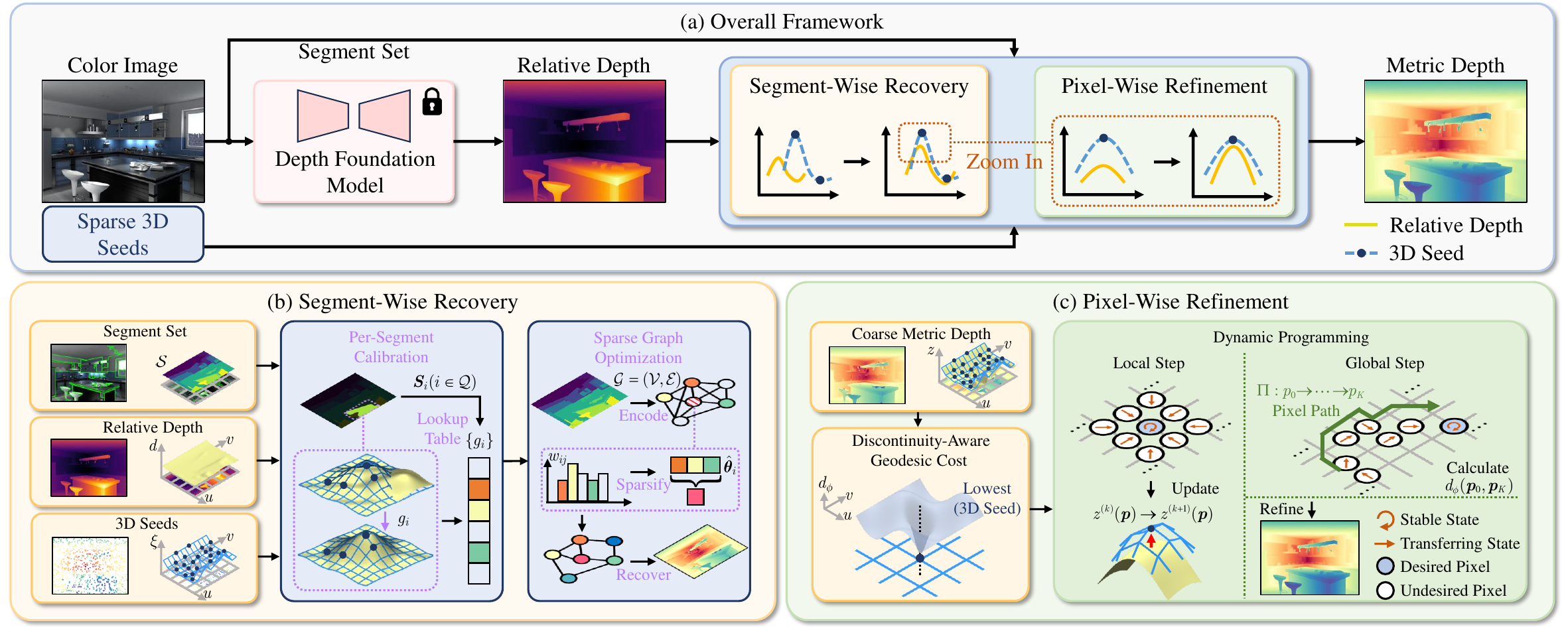}
    \vspace{-0.4em}
	\caption{MTD takes relative depth, sparse 3D seeds, and a superpixel segment set as inputs and outputs reliable metric depth. (a) In the overall framework, segment-wise recovery followed by pixel-wise refinement forms a coarse-to-fine pipeline.
    (b)
    Per-segment calibration first recovers scale for segments containing projected 3D seeds; we then propagate these calibration parameters to unseeded segments via an optimization on a segment graph.
    (c)
    Based on coarse depth, we derive pixel-wise discontinuities to guide the paths of depth propagation.
    We formulate the geodesic problem as a path-integral optimization and solve it efficiently via dynamic programming, progressing from local updates to a global solution.}
\label{fig.architectureure}
\vspace{-1.0em}
\end{figure*}

\section{Related Work}
\label{sect.related_work}

\subsection{Depth prediction}
\label{sect.related_depth_prediction}

Recent studies~\cite{geobench, depthanythingv2} build depth foundation models on large, diverse datasets, achieving strong cross-dataset generalization. Representative discriminative approaches include MiDaS~\cite{ranftl2020midas,birkl2023midasv3} and DepthAnything~\cite{depthanything,depthanythingv2}. To leverage data with varying scales, these methods typically adopt inverse-depth representations and affine-invariant losses, yielding relative depth predictions. In parallel, generative approaches~\cite{marigold,lotus,depthmaster,geowizard} repurpose diffusion models to infer dense relative depth, often improving boundary fidelity at higher inference cost. Despite strong generalization and visual quality, both categories lack absolute scale, limiting applications that require metric accuracy or spatiotemporal consistency. A complementary line of work studies metric monocular depth estimation~\cite{yin2023metric3d}, including Metric3D~\cite{yin2023metric3d, hu2024metric3dv2} and UniDepth~\cite{piccinelli2024unidepth, unidepth2}.
However, these models still face challenges in cross-domain generalization and exhibit edge-fattening issues at object boundaries compared with SoTA relative depth estimators. We instead pursue a fast, interpretable scale recovery paradigm that preserves the generalization strength of relative depth models while yielding metric depth.

\subsection{Depth Completion}
\label{sect.related_depth_completion}
Depth completion fuses the color image with sparse range measurements (\egi, LiDAR) to recover dense metric depth~\cite{lrru, cformer}. Earlier propagation-based methods~\cite{nlspn,dyspn} depend on learned affinities via additional training, making their behavior less interpretable and potentially more sensitive to domain shifts.
Recent SoTA methods, such as BP-Net~\cite{bpnet} and DMD$^{3}$C~\cite{dmd3c}, achieve strong results but tend to optimize accuracy gains on specific datasets at the expense of cross-domain robustness. Recent studies \cite{ognidc, marigolddc} have trended toward stronger generalization. OGNI-DC \cite{ognidc} introduces optimization-guided neural iterations to improve generalization. Marigold-DC~\cite{marigolddc} leverages the progressive denoising of diffusion models to refine depth completion outputs. Nevertheless, these methods~\cite{dmd3c,bpnet,marigolddc} exhibit limited flexibility and low runtime efficiency. Furthermore, achieving highly generalizable models requires collecting large-scale depth completion datasets and training for days~\cite{dmd3c}, which further increases the cost of supervised learning. In contrast, our Midas Touch approach requires no additional training, is mathematically interpretable, uses only extremely sparse 3D seed points, and flexibly accommodates multiple input modalities.

\section{Methodology}
\label{sect.method}

We present a universal, coarse-to-fine paradigm for converting relative depth to metric depth using sparse 3D seeds in a mathematically interpretable manner, as shown in Fig.~\ref{fig.architectureure}. Our approach capitalizes on the strong generalization of depth foundation models without any fine-tuning, avoiding the risk of performance degradation. Beginning with a model’s relative depth output, we first introduce a coarse, segment-wise scale recovery strategy to reduce local scale inconsistencies (Sect.~\ref{sect.segment}). We then propose a fine, pixel-wise refinement strategy by casting the task as a discontinuity-aware geodesic problem (Sect.~\ref{sect.pixel}).
Utilizing the above strategies, our method recovers accurate metric depth even when the depth foundation model is lightweight or only moderately accurate, thereby improving overall computational efficiency (Sect.~\ref{sect.framework}).

\subsection{
Segment-Wise Recovery
}
\label{sect.segment}

When converting relative depth to metric depth, the scale often varies across local regions, so a single global rescaling is insufficient. We therefore partition the image into segments. Due to the sparsity of 3D seeds, some segments receive no seed projections and lack a reliable scale. To address this, we construct a segment graph and propagate recovered scales from seeded to unseeded segments.

\subsubsection{
Per-Segment Calibration
}
\label{sect.recovery}

Given a color image $\boldsymbol{I}$ and a set of 3D seeds $\mathcal{X}$, we project $\mathcal{X}$ onto the image plane at the same resolution as $\boldsymbol{I}$. Applying superpixel segmentation~\cite{felzen, slic, mobilesam} to $\boldsymbol{I}$ produces segments $\mathcal{S}=\{\boldsymbol{S}_i\}$, where $1 \leq i \leq |\mathcal{S}| $ and $|\cdot|$ denotes set cardinality. Let $\mathcal{Q}$ index the superpixels containing projected seed priors; then $\mathcal{S} = \{\boldsymbol{S}_i\}_{i \in \mathcal{Q}} \cup \{\boldsymbol{S}_i\}_{i \notin \mathcal{Q}}$.

For each $ i \in \mathcal{Q}$, let $\mathcal{X}_i=\{\boldsymbol{X}_i^{j}\}$ denote the set of 3D seeds that projected on $\mathcal{S}_i$, where $1 \leq j \leq |\mathcal{X}_i| $. The 3D seeds originate from different sources, such as LiDAR or multi-view stereo, and each seed $\boldsymbol{X}_i^{j}$ can provide a scalar proxy $\xi_i^{j}$ that is equivalent to depth $z_i^{j}$ via a monotonic bijection (details provided in the supplementary material). Let $d_i^{j}$ be the relative depth, at the pixel corresponding to $\boldsymbol{X}_i^{j}$, predicted by a depth foundation model. We estimate a per-segment calibration function $g_i:\; d \mapsto \xi$ that maps relative depth to the scalar proxies by aligning the empirical distributions of $\{d_i^{j}\}$ and $\{\xi_i^{\,j}\}$ within $\boldsymbol{S}_i \;(i \in \mathcal{Q})$. \clc{In practice, $g_i$ can be obtained via least squares or median matching. To enable efficient parallel computation, we store the calibrated parameters of $g_i$ in a lookup table, allowing us to propagate these parameters to the uncalibrated segments in the subsequent sections.}

\subsubsection{Sparse Graph Optimization}

We extend the per-segment calibration from the supported set $\{\boldsymbol{S}_i\}_{i\in\mathcal{Q}}$ to the full segmentation $\mathcal{S}$ in a manner that is metrically faithful and spatially consistent. We encode this set $\mathcal{S}$ with a superpixel graph $\mathcal{G}=(\mathcal{V},\mathcal{E})$ whose vertices $\mathcal{V}$ correspond to segments and whose edges capture geometric proximity. For the vertex in $\mathcal{V}$, let $\boldsymbol{\theta}_i$ denote the parameters of the local calibration function $g_i$ for segment $\boldsymbol{S}_i$. $\forall i\in\mathcal{Q}$, $\hat{\boldsymbol{\theta}}_i$ is the per-segment fit obtained in Sect. \ref{sect.recovery}; $\forall i\notin\mathcal{Q}$, no seeds are available and $\boldsymbol{\theta}_i$ must be inferred from the context. For each edge $(i,j)$ in $\mathcal{E}$, edge weight $w_{ij}$ is defined by a decaying kernel of the inter-centroid distance. We also use an adaptive, median-based scale parameter that normalizes the dynamic range for numerical stability.

To reduce memory and improve computational efficiency, we sparsify $\mathcal{G}$ by retaining, for each node $v_i$, only its $N$ nearest neighbors under the distance induced by $w_{ij}$. Estimating the full set of calibration parameters $\{\boldsymbol{\theta}_i\}$ is then posed as a graph-regularized quadratic problem:
\begin{equation}\label{eq:theta}
    \min_{\{\boldsymbol{\theta}_i\}}\;
\sum_{i\in\mathcal{Q}} \,\big\|\boldsymbol{\theta}_i-\hat{\boldsymbol{\theta}}_i\big\|^2
\;+\;
\sum_{(i,j)\in\mathcal{E}} w_{ij}\,\big\|\boldsymbol{\theta}_i-\boldsymbol{\theta}_j\big\|^2.
\end{equation}
The objective encourages neighboring segments to share similar transfer parameters while remaining faithful to the per-segment anchors where available. \clc{
To solve the problem efficiently, we adopt a closed-form approximation to (\ref{eq:theta}), thereby propagating reliable calibration from $\mathcal{Q}$ into ${i\notin\mathcal{Q}}$.}

Finally, the learned graph $\mathcal{G}$ is lifted back to the image domain by assigning $g_i$ to all pixels $\boldsymbol{p}\in\mathcal{S}_i$ and applying it to the foundation model's relative depth:
$\xi(\boldsymbol{p})=g_i\big(d(\boldsymbol{p})\big)$.
This provides an estimate of metric depth, based on the known monotone bijection between $\xi$ and $z$. The segment-wise, graph-regularized transformation aligns relative depth predictions with available metric seeds and propagates a coherent, edge-aware calibration into unseeded segments.

\subsection{Pixel-Wise Refinement}
\label{sect.pixel}
Section~\ref{sect.segment} provides coarse metric depth estimations via segment-wise recovery, but per-segment matching leaves residual pixel-level errors.
A simple indication is that the real depth values on the 3D seed projections still deviate from the coarse depth. Moreover, these pixels and their neighbors often lie on the same physical surface and thus tend to share similar errors, leading to local depth inaccuracies. We therefore introduce a mathematically interpretable, pixel-wise refinement to correct these residuals.

\subsubsection{Discontinuity-Aware Geodesic Cost}
\label{sect.potential}
Our goal is to refine the coarse metric depth while discouraging wrong propagation across 3D discontinuities. We formulate the coarse metric depth $z:\Omega\subset\mathbb{R}^2\to\mathbb{R}$ as a function of image coordinates $(u,v)^{\top}$. For a pixel $\boldsymbol p\in\Omega$, we define its neighbor $ \mathcal{N}_{\boldsymbol{p}} = \{ \boldsymbol{q}:\|\boldsymbol q-\boldsymbol p\|_\infty\le 1\}$, where $\|\cdot\|_\infty$ represents the infinity norm.
\begin{proposition}[Pathwise bound on the remainder]
\label{prop.path_bound}
For any $\boldsymbol{p}$ and $\boldsymbol{q}$, define the first-order remainder
\begin{equation}\label{eq:first-order-remainder}
    R(\boldsymbol p,\boldsymbol q)
    \triangleq z(\boldsymbol q)-z(\boldsymbol p)-\tfrac12\big(\nabla z(\boldsymbol p)+\nabla z(\boldsymbol q)\big)^\top(\boldsymbol q-\boldsymbol p).
    \footnote{Compared with the common first-order remainder $r(\boldsymbol p,\boldsymbol q) = z(\boldsymbol q)-z(\boldsymbol p)-\nabla z(\boldsymbol p)^\top(\boldsymbol q-\boldsymbol p)$, the remainder in \eqref{eq:first-order-remainder} is antisymmetric, $R(\boldsymbol p,\boldsymbol q)+ R(\boldsymbol q,\boldsymbol p)=0$, hence $|R(\boldsymbol p,\boldsymbol q)|=|R(\boldsymbol q,\boldsymbol p)|$. This removes orientation bias and isolates second-order variation; in particular, $R(\boldsymbol p,\boldsymbol q)=0$ for any affine $z$.}
\end{equation}
Assume that the second partial derivatives $z_{uu}$ and $z_{vv}$ exist and are integrable.\footnote{It suffices that $z_{uu}$ and $z_{vv}$ be integrable along an axis-parallel path between $\boldsymbol{p}$ and $\boldsymbol{q}$.} Then there exists an axis-parallel polygonal path $\mathcal{L}_{\boldsymbol p\to\boldsymbol q}$ connecting $\boldsymbol p$ and $\boldsymbol q$ such that
\begin{equation}
\label{eq:path-bound}
\begin{aligned}
\bigl|R(\boldsymbol p,\boldsymbol q)\bigr|
&\le \int_{\mathcal{L}_{\boldsymbol p\to\boldsymbol q}} \phi(u,v)\,ds,\\
\phi(u,v)&\triangleq \sqrt{z_{uu}^2(u,v)+z_{vv}^2(u,v)} .
\end{aligned}
\end{equation}
\end{proposition}
\noindent
Using the Cauchy-Schwarz inequality, Proposition \ref{prop.path_bound} can be proved (details provided in the supplementary material). Proposition~\ref{prop.path_bound} guarantees the existence of at least one admissible curve; let $\mathscr{L}_{\boldsymbol p\to\boldsymbol q}$ be the family of admissible curves, which is therefore nonempty. For any $\mathcal L\in\mathscr{L}_{\boldsymbol p\to\boldsymbol q}$, the minimum accumulated quantity
\begin{equation}
\label{eq:def-dphi}
d_\phi(\boldsymbol p,\boldsymbol q)\;=\;\inf_{\mathcal L\in\mathscr{L}_{\boldsymbol p\to\boldsymbol q}} \int_{\mathcal L}\phi(u, v)\,ds
\end{equation}
is the \emph{\textit{\textbf{discontinuity-aware geodesic cost}}}, where $\inf$ denotes the infimum. $d_\phi$ is exactly the geodesic distance under the conformal Riemannian metric $\phi^2 \boldsymbol{I}_2$, where $\boldsymbol{I}_2$ denotes the identity matrix.
A rigorous proof of this statement is provided in the supplementary material.

In practice, we treat $\phi$ as a local discontinuity density and discretize the line integral in \eqref{eq:path-bound} by a Riemann-sum over single-pixel moves. For a pixel path $\Pi:\ \boldsymbol p_0\!\to\!\boldsymbol p_1\!\to\!\cdots\!\to\!\boldsymbol p_K$
and its polygonal chain $\mathcal{L}_\Pi=\bigcup_{k=0}^{K-1} \mathcal{L}_k$, the path cost can be approximated as follows:
\begin{equation}\label{eq:L_Pi}
    \int_{\mathcal{L}_\Pi}\phi\,ds\ = \sum_{k=0}^{K-1} \int_{\mathcal{L}_{k}}\phi\,ds\ \approx\ \sum_{k=0}^{K-1}\ \underbrace{\ell(\boldsymbol{p}_{k},\boldsymbol{p}_{k+1})\,\phi(\boldsymbol{p}_{k+1})}_{{W}(\boldsymbol p_k\to\boldsymbol p_{k+1})},
\end{equation}
where $\ell(\cdot,\cdot)$ is the step length. When the depth map exhibits discontinuities at object boundaries, $d_\phi$ does not violate the conditions under which \eqref{eq:path-bound}, \eqref{eq:def-dphi}, and \eqref{eq:L_Pi} hold. On the contrary, from a discretized computational perspective, such discontinuities result in a large value of $\phi$, so any curve that crosses such regions accumulates a large cost. Taking the minimum over curves yields a path-independent quantity $d_\phi$ that penalizes discontinuity crossings, thereby confining depth propagation to a reliable spatial extent.

\subsubsection{Dynamic Programming via Path Integrals}
\label{sect.dp}
Dynamic programming iteratively minimizes a cost (or energy) function using a reliable update rule, thereby converging to a stable solution. Since the geodesic cost can be approximated by \eqref{eq:L_Pi}, therefore, we rewrite it as
\begin{equation}\label{eq:dp_cost}
    d_\phi(\boldsymbol{p}_0, \boldsymbol{p}_K) \leq \inf ({W}(\boldsymbol p_{K-1}\!\to\!\boldsymbol p_{K})) + d_\phi(\boldsymbol{p}_0, \boldsymbol{p}_{K-1}),
\end{equation}
which serves as the cost function at iteration $K$. The scheme in \eqref{eq:dp_cost} establishes a correspondence between computing the line integral along a path and the cost-function optimization of dynamic programming. Furthermore, it ensures that interactions between pixels are not confined to a local neighborhood, thereby expanding the effective receptive field. We initialize the costs at the reliable projected 3D seed pixels to the minimum by default. After the dynamic-programming iterations, \eqref{eq:dp_cost} yields the smoothest discrete path among all paths from these seed pixels to $\boldsymbol{p}$ with at most $K$ moves. To update the value at a point $\boldsymbol{p}$ using $\boldsymbol{q}$ within this path, we use the following expression:
\begin{equation}\label{eq:update}
z^{(k+1)}(\boldsymbol p)
= \left(1-\tfrac{1}{k+1}\right)\, z^{(k)}(\boldsymbol p)
\;+\; \tfrac{1}{k+1}\,\hat z^{(k)}\left(\boldsymbol p \mid \boldsymbol q,\Delta\boldsymbol{p}\right),
\end{equation}
where $\hat z^{(k)}\left(\boldsymbol p \mid \boldsymbol q,\Delta\boldsymbol{p}\right) = \boldsymbol{\alpha}^{(k)}(\boldsymbol q)^\top \boldsymbol{\Psi}(\Delta\boldsymbol{p})$, $\boldsymbol{\alpha}^{(k)}(\boldsymbol q)$ denotes local coefficients estimated from data around $\boldsymbol q$ at iteration $k$, $\Delta\boldsymbol{p} = \boldsymbol{p} - \boldsymbol{q}$ denotes the one-step displacement, and $\boldsymbol{\Psi}$ is a chosen set of basis functions on the step domain. Using the harmonic step-size sequence $\tfrac{1}{k+1}$, the overall update forms a convex combination of the previous estimate and current prediction, which enhances stability while incorporating new local information.

\subsection{Computational Efficiency Improvements}
\label{sect.framework}
To improve computational efficiency and practical applicability, we reduce the parameter count of the depth foundation model via knowledge distillation. According to previous studies \cite{geobench,marigold}, high-quality datasets are crucial for distilling vision foundation models. Therefore, we leverage qualified raw data collected from both real-world and simulated environments. Inspired by the studies~\cite{mobilesam, efficientsam}, we adopt TinyViT \cite{tinyvit} and EfficientViT \cite{efficientvit} as backbones and use DepthAnythingV2 as the teacher network. We employ both feature distillation and logit distillation objectives. As shown in Sect.~\ref{sect.ablation}, despite a substantial reduction in parameters, the resulting metric depth accuracy remains comparable, and runtime efficiency improves markedly, which enables practical deployment in downstream autonomous driving and robotics applications.

\section{Experiments}
\label{sect.experiments}

\subsection{Experimental Setup}

\clb{To compare our method with existing approaches, we conduct zero-shot generalization evaluations on nuScenes~\cite{nuscenes}, DDAD~\cite{ddad}, Make3D~\cite{make3d}, DIODE~\cite{diode}, ETH3D~\cite{eth3d}, ScanNet~\cite{scannet}, VOID~\cite{void}, SUN-RGBD~\cite{sunrgbd}, HAMMER~\cite{hammer}, IBims-1~\cite{ibims}, KITTI~\cite{kitti}, and NYU-Depth V2~\cite{nyu} datasets. Detailed descriptions of the datasets’ split and statistics can be found in the supplementary material.}
The datasets used for distillation are VKITTI2~\cite{vkitti2}, Hypersim~\cite{hypersim}, TartanAir~\cite{tartanair}, and SA-1B~\cite{sam}.
The performance of depth is evaluated using several standard metrics, including root mean squared error (RMSE), mean absolute error (MAE), absolute relative error (AbsRel), squared relative error (SqRel), the accuracy metric ($\delta_i$) under thresholds of 1.25$^{i}$, and the scale-invariant error in log-scale ($\text{SI}_{\text{log}}$).

\begin{table*}[t]
\centering
\settablefont
\setlength{\tabcolsep}{8.5pt}
\caption{Quantitative comparison with SoTA depth completion methods. All methods are evaluated in a zero-shot setting. \clb{$\downarrow$ lower is better}.}
\begin{tabular}{lcccccccccc}
\toprule
\multirow{2}{*}[-0.75ex]{Method} & \multicolumn{2}{c}{nuScenes} & \multicolumn{2}{c}{DDAD} &
\multicolumn{2}{c}{Make3D} & \multicolumn{2}{c}{DIODE} & \multicolumn{2}{c}{ETH3D} \\
\cmidrule(lr){2-3}\cmidrule(lr){4-5}\cmidrule(lr){6-7}\cmidrule(lr){8-9}\cmidrule(lr){10-11}
& RMSE $\downarrow$ & MAE $\downarrow$ & RMSE $\downarrow$ & MAE $\downarrow$
& RMSE $\downarrow$ & MAE $\downarrow$ & RMSE $\downarrow$ & MAE $\downarrow$
& RMSE $\downarrow$ & MAE $\downarrow$ \\
\midrule
CFormer~\cite{cformer} \tiny\clgray{(CVPR'23)}        & 15.672 & 10.528 & 9.606 & 3.328 & 10.749 & 5.035 & 2.297 & 0.976 & 2.796 & 1.940 \\
LRRU~\cite{lrru} \tiny\clgray{(ICCV'23)}              & 13.660 & 8.472 & 9.164 & 2.738 & 13.023 & 5.893 & 2.795 & 1.947 & 3.016 & 2.337 \\
BP-Net~\cite{bpnet}  \tiny\clgray{(CVPR'24)}          & 15.092 & 10.592 & 8.903 & 2.712 & 12.034 & 5.353 & 2.724 & 1.831 & 3.365 & 2.665 \\
DMD$^3$C~\cite{dmd3c} \tiny\clgray{(CVPR'25)}         & 5.556  & 3.112 & 7.766 & 2.498 & 12.019 & 5.575 & 2.262 & 1.127 & 0.935 & 0.285 \\
PromptDA~\cite{promptda}  \tiny\clgray{(CVPR'25)}     & 9.072  & 5.325 & 8.487 & 2.891 & 12.392 & 5.803 & 2.139 & 1.287 & 1.172 & 0.479 \\
Marigold-DC~\cite{marigolddc} \tiny\clgray{(ICCV'25)} & 4.924  & 2.595 & 6.449 & 2.364 & 8.926 & 4.932 & 1.987 & 0.887 & 0.706 & 0.245 \\
\textbf{MTD (Ours)} & \textbf{4.387} & \textbf{2.177} & \textbf{5.252} & \textbf{1.834} & \textbf{8.581} & \textbf{4.776} & \textbf{1.736} & \textbf{0.761} & \textbf{0.662} & \textbf{0.177} \\
\midrule
\multirow{2}{*}[-0.75ex]{Method} & \multicolumn{2}{c}{ScanNet} & \multicolumn{2}{c}{VOID1500} &
\multicolumn{2}{c}{SUN-RGBD} & \multicolumn{2}{c}{HAMMER} & \multicolumn{2}{c}{IBims-1}\\
\cmidrule(lr){2-3}\cmidrule(lr){4-5}\cmidrule(lr){6-7}\cmidrule(lr){8-9}\cmidrule(lr){10-11}
& RMSE $\downarrow$ & MAE $\downarrow$ & RMSE $\downarrow$ & MAE $\downarrow$
& RMSE $\downarrow$ & MAE $\downarrow$ & RMSE $\downarrow$ & MAE $\downarrow$
& RMSE $\downarrow$ & MAE $\downarrow$ \\
\midrule
CFormer~\cite{cformer} \tiny\clgray{(CVPR'23)}        & 0.223 & 0.118 & 0.726 & 0.261 & 0.442 & 0.218 & 0.101 & 0.055 & 0.177 & 0.040 \\
LRRU~\cite{lrru} \tiny\clgray{(ICCV'23)}              & 1.175 & 0.773 & 0.698 & 0.232 & 1.304 & 0.808 & 1.418 & 1.137 & 0.298 & 0.107 \\
BP-Net~\cite{bpnet}  \tiny\clgray{(CVPR'24)}          & 1.326 & 1.055 & 0.704 & 0.230 & 1.327 & 1.003 & 1.620 & 1.415 & 0.302 & 0.119 \\
DMD$^3$C~\cite{dmd3c} \tiny\clgray{(CVPR'25)}         & 0.152 & 0.070 & 0.676 & 0.225 & 0.423 & 0.106 & 0.098 & 0.047 & 0.286 & 0.083 \\
PromptDA~\cite{promptda}  \tiny\clgray{(CVPR'25)}     & 0.179 & 0.072 & 0.605 & 0.191 & 0.264 & 0.095 & 0.106 & 0.070 & 0.308 & 0.122 \\
Marigold-DC~\cite{marigolddc} \tiny\clgray{(ICCV'25)} & 0.145 & 0.059 & 0.505 & 0.151 & 0.238 & 0.067 & \textbf{0.054} & 0.037 & \textbf{0.176} & \textbf{0.038} \\
\textbf{MTD (Ours)}       & \textbf{0.129} & \textbf{0.049} & \textbf{0.366} & \textbf{0.138} & \textbf{0.220} & \textbf{0.050} & 0.093 & \textbf{0.034} & 0.190 & 0.072 \\
\bottomrule
\end{tabular}\label{tab:main_results_dc}
\end{table*}

\begin{table*}[t]
\begin{center}
\settablefont
\setlength{\tabcolsep}{7.8pt}
\caption{Quantitative comparison with SoTA zero-shot monocular depth estimation methods. The upper part lists data-driven relative depth estimation methods, the middle part presents relative depth estimation methods based on diffusion models, and the lower part represents the metric depth estimation methods.}
\begin{tabular}{lccccccccccc}
\toprule
\multirow{2}{*}[-0.75ex]{Method} &
\multirow{2}{*}[-0.75ex]{\makecell[c]{with \textbf{Ours}}} &
\multicolumn{2}{c}{KITTI}   & \multicolumn{2}{c}{NYUv2} &
\multicolumn{2}{c}{ETH3D}   & \multicolumn{2}{c}{ScanNet} &
\multicolumn{2}{c}{DIODE} \\
\cmidrule(lr){3-4}\cmidrule(lr){5-6}\cmidrule(lr){7-8}\cmidrule(lr){9-10}\cmidrule(lr){11-12}
& & AbsRel $\downarrow$ & $\delta_1$ $\uparrow$
  & AbsRel $\downarrow$ & $\delta_1$ $\uparrow$
  & AbsRel $\downarrow$ & $\delta_1$ $\uparrow$
  & AbsRel $\downarrow$ & $\delta_1$ $\uparrow$
  & AbsRel $\downarrow$ & $\delta_1$ $\uparrow$ \\
\midrule
MiDaS~\cite{birkl2023midasv3}           &              & 0.183 & 0.711 & 0.095 & 0.915 & 0.190 & 0.884 & 0.099 & 0.907 & 0.266 & 0.713 \\
MiDaS~\cite{birkl2023midasv3}           & \gc{\checkmark}   & \gc{0.069} & \gc{0.929} & \gc{0.048} & \gc{0.949} & \gc{0.055} & \gc{0.944} & \gc{0.015} & \gc{0.991} & \gc{0.113} & \gc{0.864} \\
LeReS~\cite{leres}           &              & 0.149 & 0.784 & 0.090 & 0.916 & 0.171 & 0.777 & 0.091 & 0.917 & 0.271 & 0.766 \\
LeReS~\cite{leres}           & \gc{\checkmark}   & \gc{0.035} & \gc{0.974} & \gc{0.014} & \gc{0.994} & \gc{0.022} & \gc{0.984} & \gc{0.013} & \gc{0.994} & \gc{0.097} & \gc{0.885} \\
DPT~\cite{dpt}             &              & 0.111 & 0.881 & 0.091 & 0.919 & 0.115 & 0.929 & 0.084 & 0.932 & 0.269 & 0.730 \\
DPT~\cite{dpt}             & \gc{\checkmark}   & \gc{0.032} & \gc{0.976} & \gc{0.016} & \gc{0.994} & \gc{0.019} & \gc{0.985} & \gc{0.014} & \gc{0.994} & \gc{0.096} & \gc{0.887} \\
Depth Pro~\cite{depthpro}       &              & 0.077 & 0.949 & 0.044 & 0.975 & 0.060 & 0.965 & 0.042 & 0.980 & 0.321 & 0.752 \\
Depth Pro~\cite{depthpro}       & \gc{\checkmark}   & \gc{0.034} & \gc{0.975} & \gc{\textbf{0.012}} & \gc{\textbf{0.995}} & \gc{0.028} & \gc{0.974} & \gc{0.020} & \gc{0.986} & \gc{0.093} & \gc{0.889} \\
DepthAnythingV2~\cite{depthanythingv2} &              & 0.080 & 0.946 & 0.043 & 0.980 & 0.062 & 0.980 & 0.043 & 0.981 & 0.260 & 0.759 \\
DepthAnythingV2~\cite{depthanythingv2} & \gc{\checkmark}   & \gc{\textbf{0.022}} & \gc{\textbf{0.987}} & \gc{0.013} & \gc{0.995} & \gc{\textbf{0.017}} & \gc{\textbf{0.988}} & \gc{0.016} & \gc{0.991} & \gc{\textbf{0.093}} & \gc{\textbf{0.917}} \\
\midrule
Marigold~\cite{marigold}        &              & 0.099 & 0.916 & 0.055 & 0.964 & 0.065 & 0.960 & 0.064 & 0.951 & 0.308 & 0.773 \\
\clb{Marigold~\cite{marigold}}        & \gc{\checkmark}   & \gc{0.041} & \gc{0.966} & \gc{0.017} & \gc{0.991} & \gc{0.025} & \gc{0.978} & \gc{0.014} & \gc{0.991} & \gc{0.094} & \gc{0.896} \\
GeoWizard~\cite{geowizard}       &              & 0.097 & 0.921 & 0.052 & 0.966 & 0.064 & 0.961 & 0.061 & 0.953 & 0.297 & 0.792 \\
GeoWizard~\cite{geowizard}       & \gc{\checkmark}   & \gc{0.047} & \gc{0.962} & \gc{0.014} & \gc{0.994} & \gc{0.021} & \gc{0.984} & \gc{\textbf{0.013}} & \gc{\textbf{0.994}} & \gc{0.097} & \gc{0.885} \\
Lotus~\cite{lotus}            &              & 0.093 & 0.928 & 0.053 & 0.967 & 0.068 & 0.953 & 0.060 & 0.963 & 0.228 & 0.738 \\
Lotus~\cite{lotus}            & \gc{\checkmark}   & \gc{0.032} & \gc{0.975} & \gc{0.015} & \gc{0.993} & \gc{0.022} & \gc{0.985} & \gc{0.015} & \gc{0.992} & \gc{0.108} & \gc{0.868} \\
DepthMaster~\cite{depthmaster}     &              & 0.082 & 0.937 & 0.050 & 0.972 & 0.053 & 0.974 & 0.055 & 0.967 & 0.215 & 0.776 \\
DepthMaster~\cite{depthmaster}     & \gc{\checkmark}   & \gc{0.043} & \gc{0.965} & \gc{0.016} & \gc{0.993} & \gc{0.021} & \gc{0.984} & \gc{0.016} & \gc{0.992} & \gc{0.095} & \gc{0.888} \\
\midrule
Metric3Dv2~\cite{hu2024metric3dv2}      &              & 0.051 & 0.976 & 0.067 & 0.973 & 0.137 & 0.825 & 0.047 & 0.990 & 0.246 & 0.823 \\
Metric3Dv2~\cite{hu2024metric3dv2}      & \gc{\checkmark}   & \gc{0.027} & \gc{0.980} & \gc{0.014} & \gc{0.994} & \gc{0.019} & \gc{0.986} & \gc{0.015} & \gc{0.992} & \gc{0.099} & \gc{0.881} \\
\clb{UniDepthV2~\cite{unidepth2}}      &              & 0.076 & 0.952 & 0.062 & 0.970 & 0.150 & 0.865 & 0.058 & 0.975 & 0.251 & 0.795 \\
\clb{UniDepthV2~\cite{unidepth2}}      & \gc{\checkmark}   & \gc{0.032} & \gc{0.977} & \gc{0.017} & \gc{0.993} & \gc{0.023} & \gc{0.980} & \gc{0.014} & \gc{0.993} & \gc{0.096} & \gc{0.887} \\
\bottomrule
\end{tabular}\label{tab:main_results_mde}
\end{center}
\vspace{-0.5em}
\end{table*}

\begin{figure}[!t]
	\centering
	\includegraphics[width=0.49\textwidth]{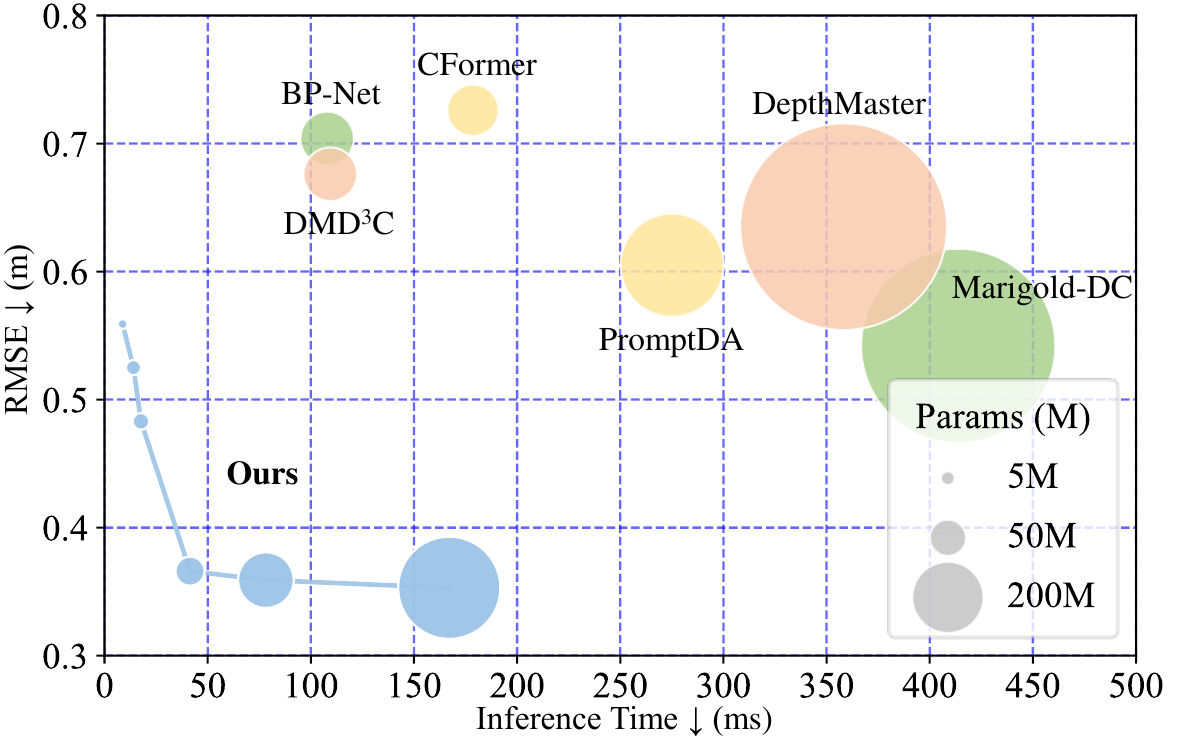}
	\caption{Evaluation results on VOID1500 at input resolution of 480$\times$640 pixels using an RTX 3090. For Ours, the left-to-right ordering corresponds to the following backbones: EfficientViT-B0, EfficientViT-B1, TinyViT, DepthAnythingV2-S, DepthAnythingV2-B, and DepthAnythingV2-L. }
\label{fig.capacity_1}
\end{figure}

\subsection{Comparison with State-of-the-Art Methods}
\label{sect.comparison_with_sota}
\noindent \textbf{Depth Completion.}
Table \ref{tab:main_results_dc} shows that our method consistently outperforms previous SoTA depth completion approaches, with improvements in terms of MAE and RMSE.
Since KITTI and NYUv2 are widely used for training, we exclude them from our zero-shot evaluation to avoid potential train-test overlap.
Outdoor scenes are generally more challenging to generalize to than indoor scenes. For instance, in datasets such as nuScenes, where point clouds are sparse and depth values are relatively high, nearly all methods struggle. Nevertheless, our approach surpasses the SoTA method Marigold-DC~\cite{marigolddc}, reducing MAE and RMSE by 0.418 and 0.537, respectively.

\noindent \textbf{Depth Prediction.}
In Table~\ref{tab:main_results_mde}, we report results for both discriminative and generative relative depth estimation methods, as well as metric depth estimation methods. For relative depth estimation, the original evaluations already require 3D data as input for global scale recovery; we preserve these input conditions and only sample 1.0\%-1.5\% of the available 3D data per scene (on KITTI, we use the official sparse data). As shown in Table~\ref{tab:main_results_mde}, MTD serves as a plug-and-play component for depth foundation models, utilizing their relative depth output and producing accurate metric depth, which consistently reduces AbsRel and increases $\delta_1$.
Moreover, Table~\ref{tab:main_results_mde} compares MTD across different depth foundation models. When paired with DepthAnythingV2, MTD achieves the best overall performance. Accordingly, we adopt DepthAnythingV2 as the primary backbone in our subsequent hardware and application experiments.

\noindent \textbf{Inference Time.} In Fig.~\ref{fig.capacity_1}, we compare our method with SoTA baselines in terms of RMSE and inference time. Benefiting from MTD’s flexibility, our framework offers greater latitude to balance accuracy and efficiency.
Notably, the dominant runtime cost lies in the front-end depth foundation model; the back end adds negligible overhead. For example, for an input image with a resolution of 480$\times$640 pixels on an RTX 3090, our back end requires only 1.9 ms, with memory usage below 1.8 GB and GPU utilization under 4\%. This underscores the importance of efficient relative depth backbones for lightweight frameworks.

\subsection{Ablation Studies}
\label{sect.ablation}
We validate the rationality and efficacy of our method through extensive ablation studies, specifically focusing on the segment-wise recovery strategy, the pixel-wise refinement strategy, the selection of the depth foundation models, and the 3D seed sparsity.

\noindent \textbf{Segment-Wise Recovery Strategy.}
As shown in Table~\ref{tab:ablation}, we evaluate both on the outdoor and indoor datasets. We first validate our per-segment calibration to determine both the proxy domain and the fitting strategy. \clb{In particular, we compare two standard alignment schemes for monocular depth estimation \cite{depthanything, marigold, monodepth2, monodepth1,dcpidepth}: median alignment and least-squares fitting. We further adopt inverse depth $z^{-1}$ as the proxy representation, yielding a measurable improvement in accuracy.} Moreover, after constructing a sparse graph over segments and optimizing it, we observe consistent improvements over the global baseline.

\noindent \textbf{Pixel-Wise Refinement Strategy.}
We solve the geodesic problem using dynamic programming and, therefore, perform ablation studies on the proposed discontinuity-aware geodesic cost and the dynamic programming jointly. As shown in Table~\ref{tab:ablation}, without our geodesic cost, standard edge extractors fail to represent discontinuities effectively. For the basis functions in (\ref{eq:update}), polynomial functions outperform B-splines. We also find that increasing $k$ in (\ref{eq:update}) expands the receptive field and improves accuracy.

\begin{table}[t]
\centering
\settablefont
\setlength{\tabcolsep}{5.5pt}
\caption{Unified ablation on KITTI (outdoor) and VOID (indoor) for segment-wise recovery and pixel-wise refinement strategy.}
\label{tab:ablation}
\begin{tabular}{llcccc}
\toprule
\multirow{2}{*}[-0.75ex]{Module} & \multirow{2}{*}[-0.75ex]{Factor} &
\multicolumn{2}{c}{KITTI} & \multicolumn{2}{c}{VOID} \\
\cmidrule(lr){3-4}\cmidrule(lr){5-6}
 & & RMSE$\downarrow$ & MAE$\downarrow$
   & RMSE$\downarrow$ & MAE$\downarrow$ \\
\midrule
\multirow{3}{*}{\makecell[l]{Per-Segment \\ Calibration}}
  & median                  &10.891 & 2.169 & 0.898  &0.358  \\
  & least squares           & 7.013 & 1.802 & 0.791  &0.307  \\
  & Domain: $z^{-1}$        & 6.782 & 1.794 & 0.614  &0.238  \\
\midrule
\multirow{2}{*}{\makecell[l]{Sparse Graph\\Optimization}}
  & global-based            & 2.521 & 0.687 &0.554  &0.169  \\
  & graph-based             & 2.232 & 0.608 &0.459  &0.150  \\
\midrule
\multirow{5}{*}{\makecell[l]{Dynamic\\ Programming}}
  & w/o $d_\phi$         & 2.618 & 0.661 &0.482  &0.158 \\
  & B-spline             & 2.112 & 0.578 &0.442  &0.147  \\
  & polynomial           & 2.049 & 0.551 &0.429  &0.145  \\
  & $k$=3                & 2.028 & 0.532 &\textbf{0.366}  &\textbf{0.138}  \\
  & $k$=5                &\textbf{1.913} & \textbf{0.498} &0.391  &0.141  \\

\bottomrule
\end{tabular}
\end{table}

\begin{figure}[!t]
	\centering
	\includegraphics[width=0.49\textwidth]{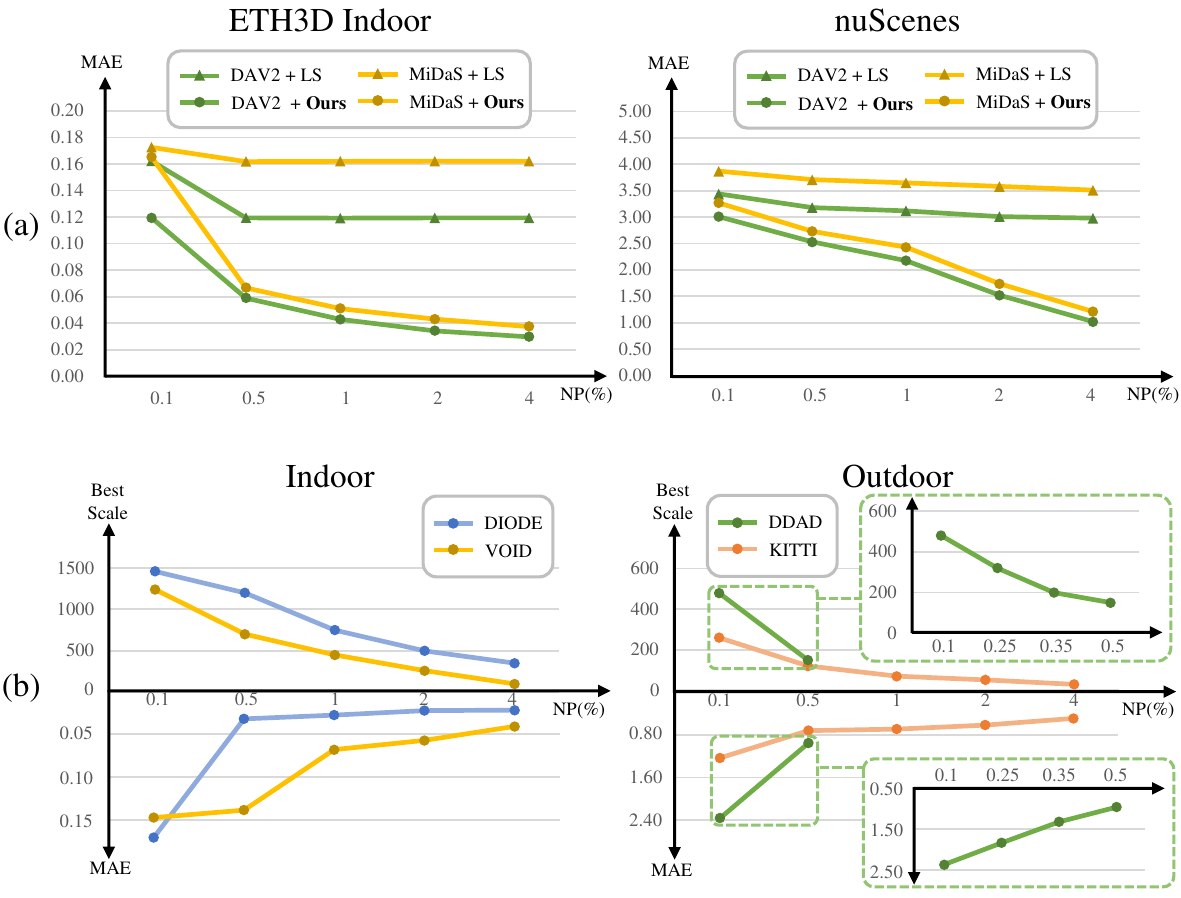}
	\caption{Ablation on 3D Seed Sparsity and Segment Scale. (a) \textbf{Effect of \emph{the number of 3D seed points} (NP)}. We compare the MAE obtained with a global least-squares baseline and with our method as NP varies. The left panel reports results on ETH3D (indoor); the right panel reports results on nuScenes (outdoor). ``LS'' represents global least-squares method, and ``DAV2'' denotes the DepthAnythingV2. (b) \textbf{Best segment scale at fixed NP}. For both indoor and outdoor datasets, we conduct a hyperparameter search over the segment scale at each fixed NP, selecting the value that minimizes MAE. }

\label{fig.np}
\end{figure}

\noindent \textbf{Foundation Model's Capacity.}
To assess our method’s practical applicability, we conduct ablation studies across different depth foundation models.
In Fig.~\ref{fig.capacity_1}, RMSE remains stable until the parameter count drops below $\sim$20M. Although RMSE increases for TinyViT, EfficientViT-B0, and EfficientViT-B1, these backbones are extremely lightweight, and their RMSE remains within an acceptable range relative to other SoTA methods. In Fig.~\ref{fig.np}(a), we evaluate with two mainstream depth foundation models, DepthAnythingV2 and MiDaS. We observe that under the global least-squares baseline, the MAE gap between DepthAnythingV2 and MiDaS is substantial; after incorporating our algorithm, this gap is markedly reduced. This demonstrates that our method can bridge the performance gaps across different depth foundation models and does not rely on high-capacity foundation models.

\begin{figure}[!t]
	\centering
	\includegraphics[width=0.49\textwidth]{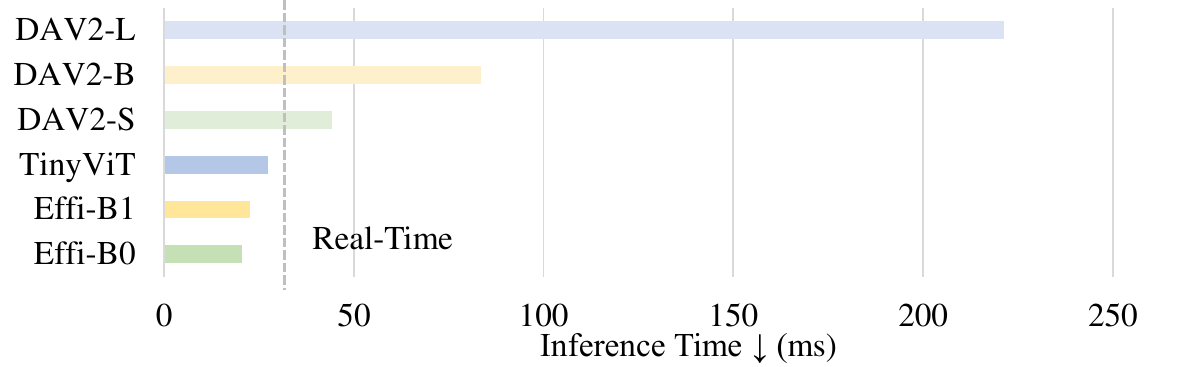}
	\caption{Inference time on embedded system (NVIDIA Jetson AGX Orin Platform) after acceleration. ``DAV2'' denotes the DepthAnythingV2, and ``Effi'' represents the EfficientViT.}
\label{fig.capacity_2}
\end{figure}

\noindent \textbf{3D Seed Sparsity and Segment Scale.}
In Fig.~\ref{fig.np}(a), we compare the MAE as the number of 3D seed points (NP) varies, using both global least squares and our method. When the 3D seeds are extremely sparse, our method degenerates to the least-squares baseline, serving as a built-in safeguard. As the point cloud becomes denser, however, our approach yields substantial MAE reductions, whereas global least squares shows little further improvement. In Fig.~\ref{fig.np}(b), for multiple datasets and NP settings, we perform a hyperparameter search over the segment-scale parameter to minimize MAE. As NP increases, the optimal segment scale consistently decreases, and the best MAE likewise drops, since additional 3D seeds enable finer metric depth detail. These trends provide practical guidance for selecting hyperparameters under different NP situations.

\begin{figure}[!t]
	\centering
	\includegraphics[width=0.49\textwidth]{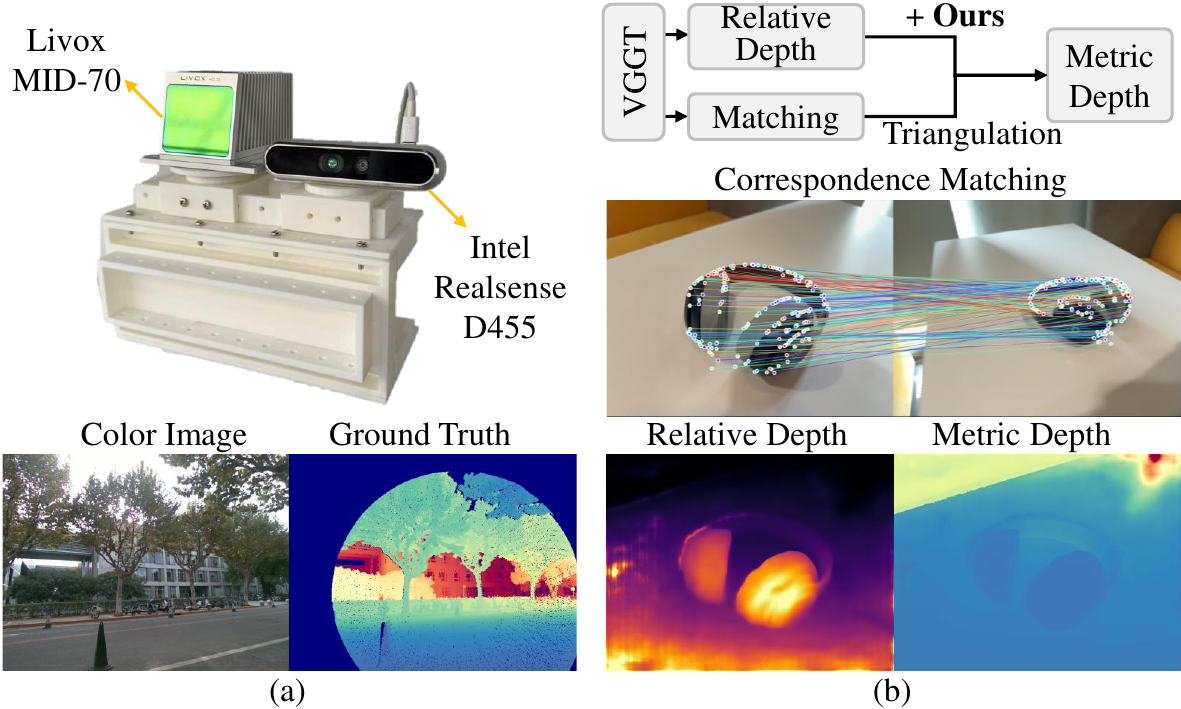}
	\caption{Evaluation setup and visualization. (a) We use a \clb{Livox MID-70 LiDAR} as ground truth to evaluate the depth rectification of the Intel RealSense D455. (b) Our method plugs into the back end of VGGT to produce metric 3D reconstructions.
    }
\label{fig.setup}
\end{figure}

\begin{figure}[!t]
	\centering
	\includegraphics[width=0.49\textwidth]{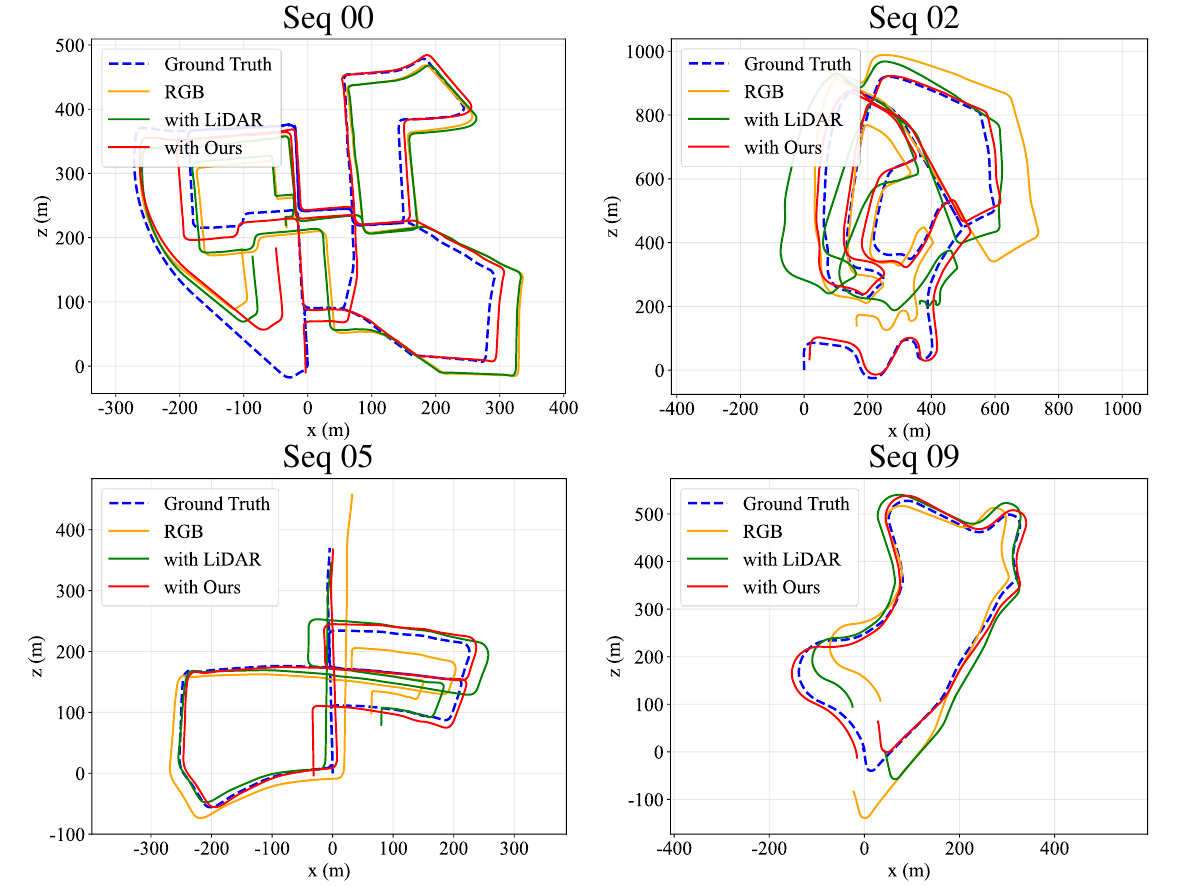}
	\caption{Qualitative results on the KITTI Odometry benchmark.}
    \vspace{-0.5em}
\label{fig.kitti_odom}
\end{figure}

\begin{table}[t]
\centering
\settablefont

\caption{\clb{Applications of MTD. (a)-(b) Experimental setups are shown in Fig.~\ref{fig.setup}. (c) We report the RMSE of the absolute trajectory error (ATE-RMSE$\downarrow$). (d) Quantitative results on KITTI-360; the evaluation metrics and the two-decimal rounding convention follow those in the study~\cite{BTS}.}}
\vspace{-0.5em}
\label{tab:applications}

\setlength{\tabcolsep}{3.63pt}
\begin{tabular}{lccccccc}
\toprule
\multicolumn{8}{c}{\textbf{(a) Depth Rectification in Commonly Used Range Camera}} \\
\midrule
\midrule
Method & Scene &
RMSE$\downarrow$ &
MAE$\downarrow$ &
AbsRel$\downarrow$ &
SqRel$\downarrow$ &
$\text{SI}_{\text{log}}\downarrow$ &
$\delta_{1}\uparrow$ \\
\midrule
Raw         & \multirow{2}{*}{Indoor} & 0.935 & 0.387  & 0.101 & 0.209 & 0.579  & 0.942  \\
Raw + \textbf{Ours}  &                         & \textbf{0.345} & \textbf{0.243}  & \textbf{0.065} & \textbf{0.032} & \textbf{0.090}  & \textbf{0.967}\\
\midrule
Raw         & \multirow{2}{*}{Outdoor} & 2.408 & 1.248  & 0.190 & 0.753 & 0.971  & 0.845  \\
Raw + \textbf{Ours}  &                          & \textbf{1.287} & \textbf{0.889}  & \textbf{0.150} & \textbf{0.289} & \textbf{0.219 }& \textbf{0.886} \\
\midrule
\end{tabular}

\setlength{\tabcolsep}{4.35pt}
\begin{tabular}{lcccccc}
\multicolumn{7}{c}{\textbf{(b) Multi-View Stereo}} \\
\midrule
\midrule
Method &
RMSE$\downarrow$ &
MAE$\downarrow$ &
AbsRel$\downarrow$ &
SqRel$\downarrow$ &
$\text{SI}_{\text{log}}\downarrow$ &
$\delta_{1}\uparrow$ \\
\midrule
MVSAnywhere\cite{mvsanywhere}        & 0.088 & 0.041  & \textbf{0.054}  & 0.011 & 0.099  & 0.957  \\
VGGT\cite{vggt} + Align       & 0.107 & 0.067 & 0.108 & 0.015 & 0.147 & 0.889 \\
VGGT\cite{vggt} + \textbf{Ours}        & \textbf{0.079} & \textbf{0.034} & 0.054 & \textbf{0.010} & \textbf{0.088} & \textbf{0.971} \\

\end{tabular}

\setlength{\tabcolsep}{11.7pt}
\begin{tabular}{lcccc}
\midrule
\multicolumn{5}{c}{\textbf{(c) SLAM on KITTI Odometry}} \\
\midrule
\midrule
Method & Seq 00 & Seq 02 & Seq 05 & Seq 06 \\
\midrule
Droid~\cite{droid}    &66.562 & 84.828 & 41.973 & 72.402 \\
Droid + LiDAR         &63.295 & 77.610 & 31.641 & 104.552 \\
Droid + \textbf{Ours} &\textbf{25.017} & \textbf{25.698} & \textbf{12.233} & \textbf{12.194}  \\
\midrule
Method & Seq 07 & Seq 08 & Seq 09 & Seq 10 \\
\midrule
Droid~\cite{droid}    &13.785 &64.552 &60.414 &11.394 \\
Droid + LiDAR         &9.867  &60.201 &42.493 &46.627 \\
Droid + \textbf{Ours} &\textbf{5.994}  &\textbf{20.738} &\textbf{23.077} &\textbf{10.871} \\
\midrule

\end{tabular}

\setlength{\tabcolsep}{10.3pt}
\begin{tabular}{lcccc}
\multicolumn{5}{c}{\textbf{(d) Occupancy Predicition}} \\
\midrule
\midrule
\multirow{2}{*}[-0.75ex]{Method}
& \multicolumn{2}{c}{Scene} & \multicolumn{2}{c}{Object} \\
\cmidrule(lr){2-3}\cmidrule(lr){4-5}
&
$\text{O}^{s}_{acc}\!\uparrow$ &
$\text{IE}^{s}_{acc}\!\uparrow$ &
$\text{O}^{o}_{acc}\!\uparrow$ &
$\text{IE}^{o}_{acc}\!\uparrow$
\\
\midrule
Monodepth2~\cite{monodepth2}            & 0.90 & N/A    & 0.69 & N/A   \\
Monodepth2 + 4m       & 0.90 & 0.59  & 0.70 & 0.53  \\
PixelNeRF~\cite{pixelNeRF}             & 0.89 & 0.62  & 0.67 & 0.53  \\
BTS~\cite{BTS}                 & 0.92 & 0.69  & 0.79 & 0.69  \\
KYN~\cite{KYN}                   & 0.92 & 0.70  & 0.79 & 0.69  \\
ViPOcc~\cite{vipocc}                & 0.93 & 0.71  & 0.79 & 0.69  \\
\clb{BTS + \textbf{Ours}}      & \textbf{0.94} & \textbf{0.72}  & \textbf{0.80} & \textbf{0.70}  \\
\bottomrule
\end{tabular}
\vspace{-1.0em}

\end{table}

\subsection{Downstream Applications}
\label{sect.application}

In Fig.~\ref{fig.capacity_2}, we further report inference time after TensorRT and multi-thread acceleration on the NVIDIA Jetson AGX Orin Platform. With our nonparametric lightweight algorithm, the embedded system achieves real-time performance. Fig.~\ref{fig.setup} and Table~\ref{tab:applications} demonstrate the effectiveness of our method across multiple downstream applications. Thanks to the plug-and-play design, our approach integrates easily into existing pipelines. Compared with raw depth from commonly used range cameras, MTD substantially improves data quality in both indoor and outdoor settings. In addition, when combined with large models such as VGGT~\cite{vggt}, our method enables multi-view stereo that surpasses the SoTA MVSAnywhere~\cite{mvsanywhere}. Specifically, VGGT outputs relative depth and correspondence matches. Utilizing the collected extrinsic, we triangulate these matches to obtain 3D seeds with known scale, which yields metric depth. Compared with a least-squares alignment, our scale-recovery strategy markedly improves the accuracy of the resulting metric depth. In SLAM experiments, augmenting LiDAR with our method significantly reduces trajectory error compared to leveraging LiDAR alone; qualitative results are shown in Fig.~\ref{fig.kitti_odom}. For occupancy prediction, following the pipeline and evaluation protocol in studies~\cite{BTS, KYN, vipocc}, we observe a consistent improvement in accuracy.

\section{Conclusion}
\label{sect.conclusion}
In this paper, we introduced a novel paradigm for recovering depth scale, comprising a segment-wise scale recovery strategy and a pixel-wise refinement strategy. Extensive experiments demonstrate strong cross-domain generalization and high accuracy of our method, and we further show that it can be flexibly integrated into a variety of downstream tasks.
As future work, we will investigate using 3D seeds only for initialization and, within a continual-learning framework, gradually eliminate this requirement.

\clearpage

\section*{Acknowledgements}
This research was supported by the National Natural Science Foundation of China under Grant 62473288, Grant 62371343, Grant 62233013, and Grant 62333017, the National Key Laboratory of Human-Machine Hybrid Augmented Intelligence, Xi'an Jiaotong University (No. HMHAI-202406), the Fundamental Research Funds for the Central Universities, NIO University Programme (NIO UP), and the Xiaomi Young Talents Program.

\appendix

\section{Ethics}
\label{Sect.ethics}
In this research, we utilize multiple datasets for depth estimation and completion, including nuScenes~\cite{nuscenes}, DDAD~\cite{ddad}, Make3D~\cite{make3d}, DIODE~\cite{diode}, ETH3D~\cite{eth3d}, ScanNet~\cite{scannet}, VOID~\cite{void}, SUN-RGBD~\cite{sunrgbd}, HAMMER~\cite{hammer}, IBims-1~\cite{ibims}, KITTI~\cite{kitti}, and NYU-Depth V2~\cite{nyu}. These datasets are used in strict accordance with their respective terms of use. While some datasets may contain images with visible faces and other personal data collected without consent, we emphasize that no processing of biometric information has occurred. We utilize the images under the CC-BY license or in a manner compatible with the Data Analysis Permission.

\section{Details of Methodology}
\label{Sect.method}

\subsection{Details on the Per-Segment Calibration}

As discussed in Sect.~\clc{3.1.1} of the main paper, we convert each 3D seed prior to a scalar cue $\xi_i^{j}$ that is equivalent to inverse depth. Concretely: (i) When metric depth $z_i^{j}$ is available (e.g., LiDAR or metric depth estimators), we set:
\begin{equation}
    \xi_i^j=\frac{\kappa}{z_s(x)+\varepsilon},
\end{equation}
where $\kappa$ is the minimum depth (note that $\kappa$ can be set to any constant, as long as the same value of
$\kappa$ is used during recovery), and $\varepsilon$ is a small constant for stability. (ii) For rectified stereo, disparity is proportional to $z^{-1}$ (up to the baseline–focal-length scale). (iii) For \clc{generic} multi-view stereo or unrectified stereo (\iei, without epipolar rectification), we perform correspondence matching and recover depth via geometric triangulation from calibrated camera poses, then we follow (i).

For $i\in \mathcal{Q}$, the calibration function $g_i$ models the per-segment transition and, in practice, can be simplified as:
\begin{equation}
    g_i(x)=\max\{a_i x+b_i,\;d_{\min}\},\qquad x\in\mathbb R_{\ge 0},
\end{equation}
where $(a_i,b_i)$ is parameters, and $d_{min}$ denotes a prescribed minimum depth.
$g_i$ can be obtained via least squares, moment matching, or quantile matching. Concretely: (i) For the least-squares situation, we adopt:
\begin{equation}
    (a_i,b_i)\;=\;\arg\min_{a,b}\sum_{x\in\mathcal S_i\cap\Omega_s}\!\big(a\,d(x)+b-\xi(x)\big)^2,
\end{equation}
where $\Omega_s$ denotes the set of projected seed pixels; (ii) for the moment matching situation, we set:
\begin{equation}
    a=\frac{\sigma_{\xi}}{\sigma_{d}},\qquad
    b=\mu_{\xi}-a\,\mu_{d},
\end{equation}
where $\sigma$ and $\mu$ denote the sample standard deviation and the sample mean, respectively. For a special case in mean scaling, we set:
\begin{equation}
    a_i=\frac{\mu_{\xi}}{\mu_{d}},\qquad b_i=0;
\end{equation}
(iii) For the quantile matching situation, we utilize the empirical quantiles $Q_p$ and the interquartile range $\mathrm{IQR}(x)=Q_{0.75}(x)-Q_{0.25}(x)$:
\begin{equation}
    a_i=\frac{{\mathrm{IQR}}(\xi)}{{\mathrm{IQR}}(d)},\qquad
    b_i=\mathrm{median}(\xi)-a_i\,\mathrm{median}(d),
\end{equation}
where $\mathrm{median}(\cdot)=Q_{0.5}(\cdot)$ denotes the median operation. For a special case in median scaling, we set:
\begin{equation}\label{eq.median}
    a_i=\frac{\mathrm{median}(\xi)}{\mathrm{median}(d)},\qquad b_i=0.
\end{equation}

\subsection{Details on the Segment Propagation and Sparse Graph Optimization}
\clc{
\noindent\textbf{Random Noise Suppression.}
For $\{\mathcal{S}_i\}_{i \notin \mathcal{Q}}$, which contain no 3D seeds, the above recovery procedure cannot be applied. Moreover, when applying $g_i$ for the calibration process on $\{\mathcal S_i\}_{i\in\mathcal Q}$, random noise becomes pronounced if the set of 3D seed priors $\mathcal{X}_i$ is extremely small. To simultaneously suppress random noise, we first construct a transfer map $\boldsymbol{T}$ from disparity $d_i$ to depth $z_i$ using data from $\{\mathcal{S}_i\}_{i\in\mathcal{Q}}$. For regions $\{\mathcal{S}_i\}_{i\notin\mathcal{Q}}$, the values in $\boldsymbol{T}$ are initialized as \clc{invalid} ($\varnothing$).

Assume $ \boldsymbol{p} $ is a 2D pixel in transformation map $\boldsymbol{T}$, and $ \mathcal{N}_{\boldsymbol{p}} $ denotes the set of neighbors of $ \boldsymbol{p} $ (including $ \boldsymbol{p} $ itself) restricted to positions where $\boldsymbol{T}$ is defined (\iei, non-$\varnothing$). $\boldsymbol{T}^{(t)}$ denotes the map at iteration $t$ of the bilateral filter. For any point $ \boldsymbol{p} $, its probability of occurrence at iteration $ t $, denoted as $ \mathbb{P}^{(t)}(\boldsymbol{p}) $, can be interpreted as the potential energy of point $ \boldsymbol{p} $ under the function $ \boldsymbol{T}^{(t)} $, i.e., $\mathbb{P}^{(t)}(\boldsymbol{p}) \propto \exp(-\boldsymbol{T}^{(t)}(\boldsymbol{p})).$
Since the state of $ \boldsymbol{p} $ is influenced by its local neighborhood $ \mathcal{N}_{\boldsymbol{p}} $, the law of total probability yields:
\begin{equation}
\mathbb{P}^{(t)}(\boldsymbol{p} \mid \mathcal{N}_{\boldsymbol{p}}) = \sum_{\boldsymbol{q} \in \mathcal{N}_{\boldsymbol{p}}} \mathbb{P}(\boldsymbol{p} \mid \boldsymbol{q}, \mathcal{N}_{\boldsymbol{p}})\, \mathbb{P}^{(t-1)}(\boldsymbol{q} \mid \mathcal{N}_{\boldsymbol{p}}),
\label{eq.total_probability}
\end{equation}
where $ \mathbb{P}(\boldsymbol{p} \mid \boldsymbol{q}) $ denotes the conditional probability of transitioning from a neighboring node $ \boldsymbol{q} $ to the central node $ \boldsymbol{p} $, and $ \mathbb{P}^{(t-1)}(\boldsymbol{q}) $ represents the prior probability of $ \boldsymbol{q} $ from the previous iteration. For each $\boldsymbol{q}\in\mathcal{N}_{\boldsymbol{p}}$, the bilateral kernel weight $ k(\boldsymbol{q}) $ is defined as:
\begin{equation}
    k(\boldsymbol{q}) = \exp\left( -\frac{\|\boldsymbol{p} - \boldsymbol{q}\|^2}{2\sigma_1^2} - \frac{ \left( \boldsymbol{I}(\boldsymbol{p}) - \boldsymbol{I}(\boldsymbol{q}) \right) ^2}{2\sigma_2^2} \right),
\end{equation}
where $\|\cdot\|$ is the Euclidean norm, $\sigma_{1}$ controls the spatial falloff, and $\sigma_{2}$ controls the color similarity. Given that the bilateral kernel $ k(\boldsymbol{q}) $ defines the influence weight of neighbor $ \boldsymbol{q} $ on the center point $ \boldsymbol{p} $, we can apply Bayes’ theorem to express:
\begin{equation}
    \mathbb{P}(\boldsymbol{p} \mid \boldsymbol{q}, \mathcal{N}_{\boldsymbol{p}}) = \frac{\mathbb{P}(\boldsymbol{q} \mid \boldsymbol{p}, \mathcal{N}_{\boldsymbol{p}})\, \mathbb{P}(\boldsymbol{p} \mid \mathcal{N}_{\boldsymbol{p}})}{\mathbb{P}(\boldsymbol{q} \mid \mathcal{N}_{\boldsymbol{p}})} \propto k(\boldsymbol{q}).
\end{equation}
Substituting this into the (\ref{eq.total_probability}) yields:
\begin{equation}
    \mathbb{P}^{(t)}(\boldsymbol{p} \mid \mathcal{N}_{\boldsymbol{p}}) = \frac{\sum_{\boldsymbol{q} \in \mathcal{N}_{\boldsymbol{p}}} k(\boldsymbol{q})\, \mathbb{P}^{(t-1)}(\boldsymbol{q} \mid \mathcal{N}_{\boldsymbol{p}})}{\sum_{\boldsymbol{q} \in \mathcal{N}_{\boldsymbol{p}}} k(\boldsymbol{q})}.
\end{equation}
Taking the logarithm of both sides leads to the filtering function:
\begin{equation}
    \boldsymbol{T}^{(t)}(\boldsymbol{p}) = -\log \frac{\sum_{\boldsymbol{q} \in \mathcal{N}_{\boldsymbol{p}}} k(\boldsymbol{q})\, \exp\left(-\boldsymbol{T}^{(t-1)}(\boldsymbol{q})\right)}{\sum_{\boldsymbol{q} \in \mathcal{N}_{\boldsymbol{p}}} k(\boldsymbol{q})}.
\end{equation}
From a geometric perspective, this filtering function effectively suppresses outliers of high-value errors through exponential weighting, thereby ensuring the stability and reliability of the transformation process.}

\noindent\textbf{Sparse Graph Optimization.}
In Sect. 3.1.2 of the main paper, we introduce a sparse graph optimization. We first model the $\{\mathcal{S}_i\}$ into a graph $\mathcal{G}$. The vertex of $\mathcal{G}$ stores the calibration parameters, and the edge of $\mathcal{G}$ is the spatial distance of the center in  $\{\mathcal{S}_i\}$. We then sparsify the graph by sorting the edge weights of $\mathcal{G}$. We preserve the $N$-minimum distance of each vertex.
Let $\boldsymbol{c}_i$ denote the centroid of $\boldsymbol{S}_i$. The $w_{ij}$ in Sect.~3.1.2 can be computed as follows:
$ w_{ij}\propto\exp\left(-{\|\boldsymbol{c}_i-\boldsymbol{c}_j\|}/{\tau}\right)$,
where $\|\cdot\|$ is the Euclidean norm, and $\tau$ is the adaptive, median-based scale parameter for numerical stability.

\subsection{Proof of Proposition 1 in the Main Paper}

\noindent\textbf{Diagonal case.}
Let $\boldsymbol p=(u_0,v_0)^{\top}$ and $\boldsymbol q=(u_1,v_1)^{\top}$.
Consider the two axis-aligned polyline paths between $\boldsymbol p $ and $\boldsymbol q $. Let $\mathcal L_1$ be the axis-aligned path $\boldsymbol p\!\to\!(u_1,v_0)^{\top}\!\to\!\boldsymbol q$ (horizontal then vertical), and
$\mathcal L_2$ the path $\boldsymbol p\!\to\!(u_0,v_1)^{\top}\!\to\!\boldsymbol q$ (vertical then horizontal). Define the symmetric first-order remainder as follows:
\begin{equation}
\begin{aligned}
R(\boldsymbol p,\boldsymbol q)
:= z(\boldsymbol q)-z(\boldsymbol p)
-\tfrac12\big(\nabla z(\boldsymbol p)+\nabla z(\boldsymbol q)\big)^\top(\boldsymbol q-\boldsymbol p).
\end{aligned}
\end{equation}
Our goal is to choose one path such that $|R(\boldsymbol p,\boldsymbol q)|$ is bounded by a path integral $\int\phi\,ds$, where $\phi(u,v)=\sqrt{z_{uu}^2(u,v)+z_{vv}^2(u,v)}$ and $ds = \sqrt{du^2+dv^2}$.

We first rewrite $R(\boldsymbol p,\boldsymbol q)$ as the average of the remainders associated with the two candidate paths. Set $\Delta u=u_1-u_0$, $\Delta v=v_1-v_0$.
For each one-dimensional axis-aligned path (horizontal paths at fixed $v$, vertical paths at fixed $u$), we apply the integral form of the trapezoidal rule error for a single variable and then integrate by parts along that path. This yields the following two equations:
\begin{equation}\label{eq:hv}
\begin{aligned}
R_{H\to V}
&\coloneqq z(\boldsymbol q)-z(\boldsymbol p) \\
& \;\;\quad -\big(\Delta u\,z_u(\boldsymbol p)+\Delta v\,z_v(\boldsymbol q)\big) \\
&\;\;=\int_{u_0}^{u_1}(u_1-u)\,z_{uu}(u,v_0)\,du \\
&\;\;\quad -\int_{v_0}^{v_1}(v-v_0)\,z_{vv}(u_1,v)\,dv,
\end{aligned}
\end{equation}
\begin{equation}\label{eq:vh}
\begin{aligned}
R_{V\to H}
&\coloneqq z(\boldsymbol q)-z(\boldsymbol p) \\
& \;\;\quad -\big(\Delta v\,z_v(\boldsymbol p)+\Delta u\,z_u(\boldsymbol q)\big) \\
&\;\;=-\int_{u_0}^{u_1}(u-u_0)\,z_{uu}(u,v_1)\,du \\
&\;\;\quad +\int_{v_0}^{v_1}(v_1-v)\,z_{vv}(u_0,v)\,dv.
\end{aligned}
\end{equation}
These two formulas can be proved using a one-dimensional line integral. By definition and a simple rearrangement, we obtain the following equation:
\begin{equation}\label{eq:two}
\begin{aligned}
R(\boldsymbol p,\boldsymbol q)
=\tfrac12\,R_{H\to V}+\tfrac12\,R_{V\to H}.
\end{aligned}
\end{equation}
From \eqref{eq:hv} and the triangle inequality,
\begin{equation}
\begin{aligned}
|R_{H\to V}|
&\le \int_{u_0}^{u_1}(u_1-u)\,|z_{uu}(u,v_0)|\,du \\
&\quad +\int_{v_0}^{v_1}(v-v_0)\,|z_{vv}(u_1,v)|\,dv.
\end{aligned}
\end{equation}
Since $0\le (u_1-u)\le |u_1-u_0|=|\Delta u|$ and $0\le(v-v_0)\le |\Delta v|$, we obtain the following equation:
\begin{equation}
\begin{aligned}
|R_{H\to V}|
&\le |\Delta u|\int_{u_0}^{u_1}|z_{uu}(u,v_0)|\,du \\
&\quad +|\Delta v|\int_{v_0}^{v_1}|z_{vv}(u_1,v)|\,dv.
\end{aligned}
\end{equation}
On the horizontal path $ds=|du|$ and $|z_{uu}|\le \phi$; on the vertical path $ds=|dv|$ and $|z_{vv}|\le \phi$. Therefore, we obtain the following equation:
\begin{equation}\label{eq.1_min}
|R_{H\to V}|\ \le\ \int_{\mathcal L_{H\to V}}\phi(u,v)\,ds.
\end{equation}
From \eqref{eq:vh}, we can obtain the following expression:
\begin{equation}\label{eq.2_min}
 |R_{V\to H}|\ \le\ \int_{\mathcal L_{V\to H}}\phi(u,v)\,ds.
\end{equation}
Take absolute values in \eqref{eq:two} and apply \eqref{eq.1_min}-\eqref{eq.2_min}:
\begin{equation}
\begin{aligned}
|R|
&\le \tfrac12\int_{\mathcal L_{H\to V}}\phi\,ds \;\;+\;\;\tfrac12\int_{\mathcal L_{V\to H}}\phi\,ds \\
&\le \max\{\int_{\mathcal L_{H\to V}}\phi\,ds,\ \int_{\mathcal L_{V\to H}}\phi\,ds\}.
\end{aligned}
\end{equation}
Hence there exists $\mathcal L\in\{\mathcal L_{H\to V},\mathcal L_{V\to H}\}$ such that
\begin{equation}
\big|R(\boldsymbol p,\boldsymbol q)\big|\ \le\ \int_{\mathcal L}\phi(u,v)\,ds.
\end{equation}
This proves the claimed pathwise existence bound (the Cauchy-Schwarz inequality can also be employed to obtain the same result).

\noindent\textbf{Purely horizontal step.}
We derive the following equation:
\begin{equation}\label{eq:hor-identity}
\begin{aligned}
R^{\rm(hor)}
= \int_{u_0}^{u_1}\Big(\tfrac{u_0+u_1}{2}-u\Big)\,z_{uu}(u,v_0)\,du.
\end{aligned}
\end{equation}
We then derive the following equation:
\begin{equation}\label{eq:hor-final}
\begin{aligned}
\big|R^{\rm(hor)}\big|
&\le \int_{u_0}^{u_1}\Big|\tfrac{u_0+u_1}{2}-u\Big|\,|z_{uu}(u,v_0)|\,du \\
&\le \int_{u_0}^{u_1}|z_{uu}(u,v_0)|\,du \\
&\le\; \int_{\mathcal L_{\rm hor}}\phi\,ds,
\end{aligned}
\end{equation}
where $\mathcal L_{\rm hor}$ is the single horizontal axis-aligned path (so $ds=|du|$ and $|z_{uu}|\le\phi$ along it).

\noindent\textbf{Purely vertical step.}
Similarly, we can derive the following equation:
\begin{equation}\label{eq:ver-identity}
\begin{aligned}
R^{\rm(ver)}
= \int_{v_0}^{v_1}\Big(\tfrac{v_0+v_1}{2}-v\Big)\,z_{vv}(u_0,v)\,dv,
\end{aligned}
\end{equation}
and
\begin{equation}\label{eq:ver-final}
\begin{aligned}
\big|R^{\rm(ver)}\big|
&\le \int_{v_0}^{v_1}|z_{vv}(u_0,v)|\,dv
\le\; \int_{\mathcal L_{\rm ver}}\phi\,ds,
\end{aligned}
\end{equation}
where $\mathcal L_{\rm ver}$ is the single vertical axis-aligned path (so $ds=|dv|$ and $|z_{vv}|\le\phi$ along it).

\subsection{Proof of Geodesic Cost}
Define the weighted path length and the induced geodesic cost as follows:
\begin{equation}
    L_\phi(\gamma)=\int_\gamma \phi\,ds,\qquad
    d_\phi(\boldsymbol{p},\boldsymbol{q})=\inf_{\gamma:\,\boldsymbol{p}\to \boldsymbol{q}} L_\phi(\gamma),
\end{equation}
where for a piecewise $C^1$ curve $\gamma:[0,1]\to\Omega$, we have $ds=\|\dot\gamma(t)\|_2\,dt$. Endow $\Omega\subset\mathbb{R}^2$ with the conformal Riemannian metric $g=\phi^2 \boldsymbol{I}_2$, where $\boldsymbol{I}_2$ is the identity matrix.
Then the Riemannian length of $\gamma$ is:
\begin{equation}
\begin{aligned}
L_g(\gamma)
&= \int_0^1 \sqrt{\,\dot\gamma(t)^\top g_{\gamma(t)}\,\dot\gamma(t)\,}\,dt \\
&= \int_0^1 \phi(\gamma(t))\,\|\dot\gamma(t)\|_2\,dt \\
&= \int_\gamma \phi\,ds .
\end{aligned}
\end{equation}
Hence,
\begin{equation}
L_\phi(\gamma)=L_g(\gamma)\,\quad\Rightarrow\quad
\,d_\phi(\boldsymbol{p},\boldsymbol{q})=d_g(\boldsymbol{p},\boldsymbol{q})\,.
\end{equation}
The path length formulation exactly is the geodesic distance under the conformal metric $g=\phi^2 \boldsymbol{I}_2$ (identical values and the same minimizing curves up to reparameterization).

\subsection{Details on the Dynamic Programming}
To update the value at a point $ \boldsymbol{p} $ using information from its neighboring point $ \boldsymbol{q} $ in the dynamic programming process, we use the following expansion in Sect.~3.2.2:
\begin{equation}\label{eq:update}
z^{(k+1)}(\boldsymbol p)
= \left(1-\tfrac{1}{k+1}\right)\, z^{(k)}(\boldsymbol p)
\;+\; \tfrac{1}{k+1}\,\hat z^{(k)}\left(\boldsymbol p \mid \boldsymbol q,\Delta\boldsymbol{p}\right).
\end{equation}
Under a polynomial expansion, the above expression can be reformulated as the following equation:
\begin{equation}\label{eq.all}
\begin{aligned}
z^{(k+1)}(\boldsymbol p)
= &\ \left(1 - \frac{1}{k+1}\right) z^{(k)}(\boldsymbol p)
+ \frac{1}{k+1} \bigg(
    z^{(k)}(\boldsymbol q)  \\
    &\ + \nabla z^{(k)}(\boldsymbol q)^\top \Delta \boldsymbol{h}
    + \frac{1}{2} \Delta \boldsymbol{h}^\top \boldsymbol H_z^{(k)}(\boldsymbol q) \Delta \boldsymbol{h} + \cdots
\bigg),
\end{aligned}
\end{equation}
where the displacement vector $\Delta \boldsymbol{h} =  \boldsymbol{p}- \boldsymbol{q}$. \(\nabla z^{(k)}(\boldsymbol q)\) is the gradient of \(z^{(k)}\) at point \(\boldsymbol q\), and \(\boldsymbol H_z^{(k)}(\boldsymbol q)\) is the Hessian matrix of \(z^{(k)}\) at \(\boldsymbol q\), given by:
\begin{equation}
\boldsymbol H_z^{(k)}(\boldsymbol q) = \begin{bmatrix}
z_{xx}^{(k)}(\boldsymbol q) & z_{xy}^{(k)}(\boldsymbol q) \\
z_{xy}^{(k)}(\boldsymbol q) & z_{yy}^{(k)}(\boldsymbol q)
\end{bmatrix},
\end{equation}
where $ z_{xx} $, $ z_{yy} $, and $ z_{xy} $ denote the second partial derivatives of $z$ with respect to $x$, $y$, and the mixed partial derivative with respect to $x$ and $y$, respectively.

For (\ref{eq.all}), we use the step sizes $\tfrac{1}{k+1}$, which satisfy the following conditions:
\begin{equation}
    \sum_{k=1}^\infty \frac{1}{k} = \infty, \quad \sum_{k=1}^\infty \left(\frac{1}{k}\right)^2 < \infty.
\end{equation}
Intuitively, the first condition prevents the updates from vanishing too quickly, so new information continues to influence the estimate; the second mitigates the cumulative effect of noise.
We now make the convergence analysis fully two-dimensional by treating the Taylor remainder $R^{(k)}_2(\boldsymbol p,\boldsymbol q)$ rigorously.
Assume that each iterate $z^{(k)}:\mathbb R^2\to\mathbb R$ is three-times continuously differentiable
on an open set containing the line path between $\boldsymbol p$ and $\boldsymbol q$, and that the
third-derivative tensors are uniformly bounded along these paths. Specifically, there exists $ M>0 $ such that
\begin{equation}
\sup_{\boldsymbol \xi} \bigl\|\nabla^3 z^{(k)}(\boldsymbol \xi)\bigr\|_{\mathrm{op}} \le M, \quad \forall\;k.
\end{equation}
Here, $\|\cdot\|_{\mathrm{op}}$ denotes the operator norm. For each iteration, the second-order Taylor remainder can be written in the form:
\begin{equation}
\epsilon_{k+1} \coloneqq R^{(k)}_2(\boldsymbol p,\boldsymbol q)
=\tfrac{1}{6} \nabla^3 z^{(k)}(\boldsymbol \xi_{k+1})\bigl[\Delta\boldsymbol h,\Delta\boldsymbol h,\Delta\boldsymbol h\bigr],
\end{equation}
where $\boldsymbol \xi_{k+1}$ is in the line path between $\boldsymbol p$ and $\boldsymbol q$. This implies the bound:
\begin{equation}
|\epsilon_{k+1}|\le \tfrac{M}{6}\,\|\Delta\boldsymbol h\|^3.
\end{equation}
Therefore, the step size $\Delta \boldsymbol h$ should be controlled.

\begin{figure*}[!t]
	\centering
	\includegraphics[width=0.99\textwidth]{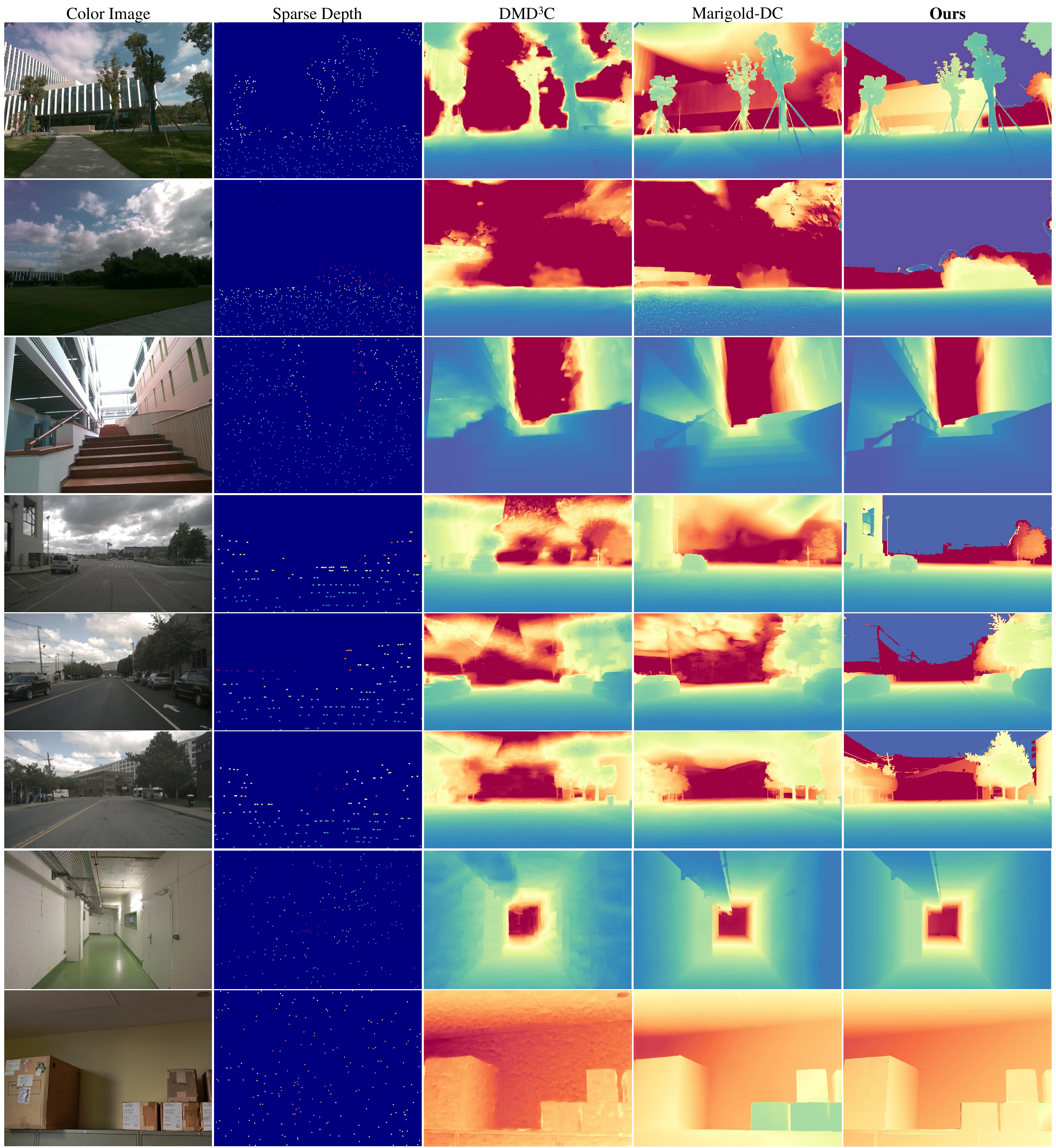}
	\caption{Qualitative comparisons with SoTA depth completion methods.}
\label{fig.qual}
\end{figure*}

\begin{figure}[!t]
	\centering
	\includegraphics[width=0.49\textwidth]{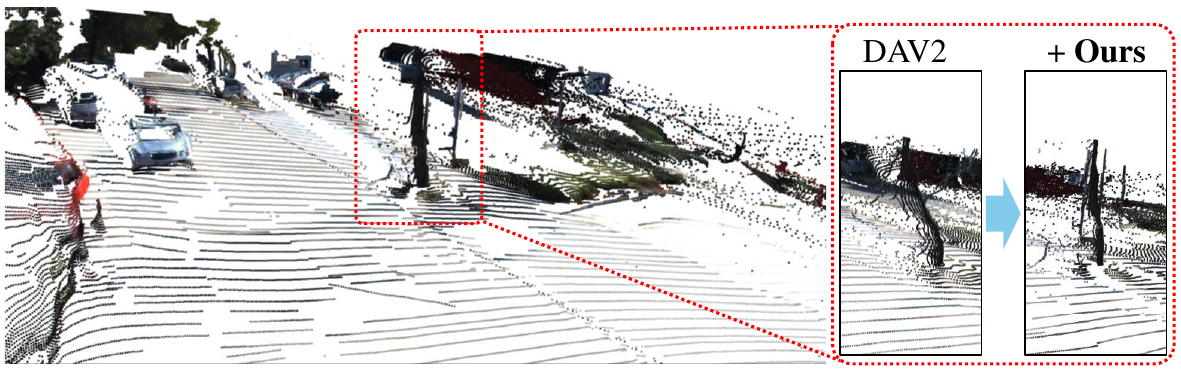}
	\caption{Illustration of reducing local scale inconsistencies in our method. ``DAV2'' denotes the DepthAnythingV2, which exhibits noticeable local scale inconsistencies around the tree trunks.}
\label{fig.dc_illu}
\vspace{1.0em}
\end{figure}

 \begin{figure}[!t]
	\centering
	\includegraphics[width=0.49\textwidth]{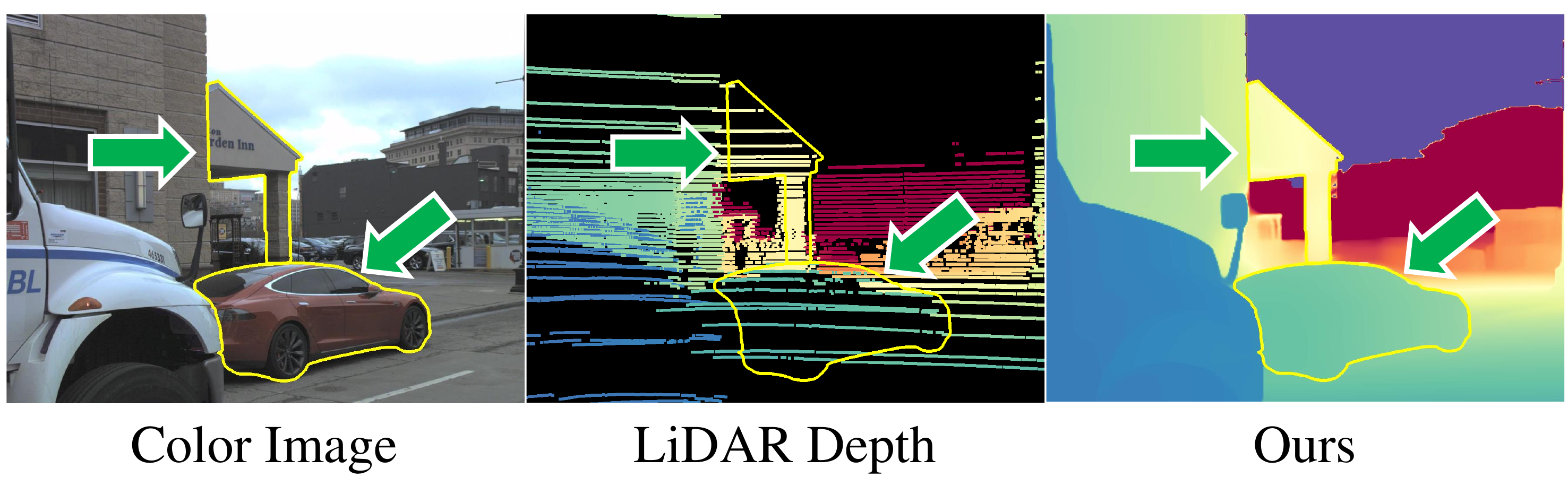}
        \vspace{-1.5em}
        \settablefont
	\caption{RGB-LiDAR misalignment on Argoverse 2.}
\label{fig.supp_rebuttal}
\end{figure}

\subsection{Details on the Knowledge Distillation}
We leverage high-quality raw data collected from both real-world and virtual scenarios. Our objective is to develop a lightweight encoder capable of robust generalization for applications. To achieve this, we employ DepthAnythingV2~\cite{depthanythingv2} as a teacher network to distill knowledge into our depth foundation models. As shown in previous work~\cite{geobench}, virtual datasets offer advantages such as high quality and precise synchronization. Moreover, high-quality datasets tend to provide more reliable supervision for pretrained models~\cite{geobench}. Therefore, we utilize both high-quality virtual and real-world datasets, including VKITTI2~\cite{vkitti2}, Hypersim~\cite{hypersim}, TartanAir~\cite{tartanair}, and SA-1B~\cite{sam}.
During training, both the depth foundation models and the DPT decoder~\cite{dpt} are trained using a combination of real-world and virtual datasets. To facilitate training, the decoder is constrained by pseudo labels generated by DepthAnythingV2.
Assume the input image is denoted as $\boldsymbol{I}$.
Our teacher features from the DepthAnythingV2 image encoder can be written as
$ f^{\text{da}}(\boldsymbol{I})$,
the output from DepthAnythingV2 DPT head is $ \boldsymbol D^{\text{da}} = h^{\text{da}}(f^{\text{da}}(\boldsymbol{I})) $. The output from the lightweight depth foundation model's encoder is $ f^{\text{tiny}}(\boldsymbol{I})$, and the lightweight DPT decoder output is $ \boldsymbol D^{\text{tiny}} = h^{\text{tiny}}(f^{\text{tiny}}(\boldsymbol{I})) $. The overall loss is defined as the sum of the logit distillation loss \(\mathcal{L}_{l}\) and the feature distillation \(\mathcal{L}_{f}\).
The logit distillation loss between the DepthAnythingV2~\cite{depthanythingv2} and our lightweight DPT decoder~\cite{dpt} can be defined as follows:
\begin{equation}
\begin{aligned}
\mathcal{L}_{l} = &\frac{1}{N} \sum_{i=1}^{N} \left| \rho(\boldsymbol D^{\text{da}}) - \rho(\boldsymbol D^{\text{tiny}}) \right|  \\
&+ \alpha \cdot \frac{1}{N} \sum_{i=1}^{N}
\Big( \big| \rho (\nabla_x \boldsymbol D^{\text{da}}) - \rho(\nabla_x \boldsymbol D^{\text{tiny}}) \big| \Big)\\
&+ \alpha \cdot \frac{1}{N} \sum_{i=1}^{N} \Big(\big| \rho (\nabla_y \boldsymbol D^{\text{da}}) - \rho(\nabla_y \boldsymbol D^{\text{tiny}}) \big| \Big),
\end{aligned}
\end{equation}
where $N$ is the number of pixels, $|\cdot|$ denotes a $\ell_1$ norm, $\alpha$ is set as in~\cite{depthanythingv2}, and the function $ \rho $ is the scaled and shifted operation. The feature distillation loss can be defined as follows:
\begin{equation}
\mathcal{L}_{f} = \frac{1}{M\cdot N} \sum_{m=1}^{M} \sum_{j=1}^{N} \left\| f^{\text{da}}(\boldsymbol{I}) - f^{\text{tiny}}(\boldsymbol{I}) \right\|^2,
\end{equation}
where $M$ represents the number of multi-scale features.

\begin{table*}[t]
\centering
\settablefont
\setlength{\tabcolsep}{8.5pt}
\caption{Quantitative comparison with SoTA depth completion methods. All methods are evaluated in a zero-shot setting.}
\begin{tabular}{lcccccccccc}
\toprule
\multirow{2}{*}[-0.75ex]{Method} & \multicolumn{2}{c}{nuScenes} & \multicolumn{2}{c}{VOID1500} &
\multicolumn{2}{c}{VOID500} & \multicolumn{2}{c}{VOID150} & \multicolumn{2}{c}{IBims-1} \\
\cmidrule(lr){2-3}\cmidrule(lr){4-5}\cmidrule(lr){6-7}\cmidrule(lr){8-9}\cmidrule(lr){10-11}
& RMSE $\downarrow$ & MAE $\downarrow$ & RMSE $\downarrow$ & MAE $\downarrow$
& RMSE $\downarrow$ & MAE $\downarrow$ & RMSE $\downarrow$ & MAE $\downarrow$
& RMSE $\downarrow$ & MAE $\downarrow$ \\
\midrule
DepthAnythingV2~\cite{depthanythingv2}              & 5.303 & 3.163 & 0.605 & 0.209 & 0.582 & 0.209 & 0.644 & 0.230 & 0.305 & 0.132 \\
Depth Pro~\cite{depthpro}                    & 6.232 & 3.657 & 0.734 & 0.385 & 0.697 & 0.373 & 0.758 & 0.392 & 0.332 & 0.148 \\
Marigold~\cite{marigold}                     & 5.459 & 3.271 & 0.630 & 0.240 & 0.607 & 0.241 & 0.673 & 0.263 & 0.306 & 0.138 \\
DPromting~\cite{depthprompting}                    & 13.981 & 9.182 & 0.779 & 0.373 & 0.754 & 0.373 & 0.820 & 0.398 & 0.297 & 0.102 \\
BP-Net~\cite{bpnet}          & 15.092 & 10.592 & 0.738 & 0.268 & 0.790 & 0.369 & 0.934 & 0.470 & 0.302 & 0.119 \\
DMD$^3$C~\cite{dmd3c}        & 5.556  & 3.112  & 0.676 & 0.225 & 0.736 & 0.275 & 0.762 & 0.297 & 0.286 & 0.083 \\
G2-Monodepth~\cite{g2_monodepth}                 & 8.921  & 4.587 & 0.568 & 0.159 & 0.574 & 0.182 & 0.691 & 0.247 & 0.267 & 0.078 \\
Marigold-DC~\cite{marigolddc}& 4.924  & 2.595  & 0.505 & 0.151 & 0.535 & 0.158 & 0.622 & \textbf{0.194} & 0.176 & 0.038 \\
\textbf{MTD (Ours)}          & \textbf{4.387} & \textbf{2.177} & \textbf{0.366} & \textbf{0.138} & \textbf{0.522} & \textbf{0.157} & \textbf{0.615} & 0.217 & 0.190 & 0.072 \\
\bottomrule
\end{tabular}\label{tab:supp_dc}

\end{table*}

\begin{table*}[t]
\begin{center}
\settablefont
\setlength{\tabcolsep}{9.4pt}
\caption{Quantitative comparison with SoTA zero-shot depth estimation methods. We utilize the metric depth estimator to generate the 3D seeds for depth estimation.}
\begin{tabular}{lcccccccccc}
\toprule
\multirow{2}{*}[-0.75ex]{Method} &
\multicolumn{2}{c}{KITTI}   & \multicolumn{2}{c}{NYUv2} &
\multicolumn{2}{c}{ETH3D}   & \multicolumn{2}{c}{ScanNet} &
\multicolumn{2}{c}{DIODE} \\
\cmidrule(lr){2-3}\cmidrule(lr){4-5}\cmidrule(lr){6-7}\cmidrule(lr){8-9}\cmidrule(lr){10-11}
& AbsRel $\downarrow$ & $\delta_1$ $\uparrow$
  & AbsRel $\downarrow$ & $\delta_1$ $\uparrow$
  & AbsRel $\downarrow$ & $\delta_1$ $\uparrow$
  & AbsRel $\downarrow$ & $\delta_1$ $\uparrow$
  & AbsRel $\downarrow$ & $\delta_1$ $\uparrow$ \\
\midrule
DiverseDepth~\cite{diversedepth}       & 0.190 & 0.704 & 0.117 & 0.875 & 0.228 & 0.694 & 0.109 & 0.882 & 0.376 & 0.631 \\
MiDaS~\cite{birkl2023midasv3}          & 0.183 & 0.711 & 0.095 & 0.915 & 0.190 & 0.884 & 0.099 & 0.907 & 0.266 & 0.713 \\
LeReS~\cite{leres}                     & 0.149 & 0.784 & 0.090 & 0.916 & 0.171 & 0.777 & 0.091 & 0.917 & 0.271 & 0.766 \\
Omnidata~\cite{omnidata}                               & 0.149 & 0.835 & 0.074 & 0.945 & 0.166 & 0.778 & 0.075 & 0.936 & 0.339 & 0.742 \\
HDN~\cite{HDN}                                    & 0.115 & 0.867 & 0.069 & 0.948 & 0.121 & 0.833 & 0.080 & 0.939 & 0.246 & 0.780 \\
DPT~\cite{dpt}                         & 0.111 & 0.881 & 0.091 & 0.919 & 0.115 & 0.929 & 0.084 & 0.932 & 0.269 & 0.730 \\
Depth Pro~\cite{depthpro}              & 0.077 & 0.949 & 0.044 & 0.975 & 0.060 & 0.965 & \textbf{0.042} & 0.980 & 0.321 & 0.752 \\
DepthAnythingV2~\cite{depthanythingv2} & 0.080 & 0.946 & 0.043 & 0.980 & 0.062 & 0.980 & 0.043 & \textbf{0.981} & 0.260 & 0.759 \\
Marigold~\cite{marigold}               & 0.099 & 0.916 & 0.055 & 0.964 & 0.065 & 0.960 & 0.064 & 0.951 & 0.308 & 0.773 \\
GeoWizard~\cite{geowizard}             & 0.097 & 0.921 & 0.052 & 0.966 & 0.064 & 0.961 & 0.061 & 0.953 & 0.297 & 0.792 \\
Lotus~\cite{lotus}                     & 0.093 & 0.928 & 0.053 & 0.967 & 0.068 & 0.953 & 0.060 & 0.963 & 0.228 & 0.738 \\
DepthMaster~\cite{depthmaster}         & 0.082 & 0.937 & 0.050 & 0.972 & 0.053 & 0.974 & 0.055 & 0.967 & 0.215 & 0.776 \\
\textbf{Ours}                          & \textbf{0.075} & \textbf{0.953} & \textbf{0.038} & \textbf{0.980} & \textbf{0.051} & \textbf{0.981} & 0.044 & 0.978 & \textbf{0.207} & \textbf{0.816} \\
\bottomrule
\end{tabular}\label{tab:supp_mde}
\end{center}
\vspace{-0.5em}
\end{table*}

\begin{table}[!t]
\centering
\settablefont
\setlength{\tabcolsep}{1.7pt}
\caption{Zero-shot evaluation.
$^{*}$ NLSPN's KITTI-DC results from VPP4DC (as in Marigold-DC).
Prop-Time is benchmarked on an RTX 3090 with 480$\times$640 inputs. For SPN-based methods, we report ``guidance generation + propagation'' time; for ours, we report the full pixel-wise refinement time, as guidance and propagation are performed jointly.
For completeness, the total back-end runtime (segment-wise recovery + pixel-wise refinement) is 1.9 ms.
}
\label{tab:spn}
\begin{tabular}{l cc cc cc c}
\toprule
\multirow{2}{*}[-0.75ex]{Method}
& \multicolumn{2}{c}{nuScenes}
& \multicolumn{2}{c}{VOID}
& \multicolumn{2}{c}{KITTI-DC}
& \multirow{2}{*}[-0.75ex]{\shortstack[c]{Prop-\\Time (ms)}} \\
\cmidrule(lr){2-3}\cmidrule(lr){4-5}\cmidrule(lr){6-7}
& RMSE $\downarrow$ & MAE $\downarrow$
& RMSE $\downarrow$ & MAE $\downarrow$
& RMSE $\downarrow$ & MAE $\downarrow$
& \\
\midrule
CSPN~\cite{spn1}          & 15.871 & 10.792 & 1.571 & 0.548 & -     & -     & \;\,46.5+22.1 \\
NLSPN$^{*}$~\cite{nlspn}   & 13.630 & 8.809  & 1.353 & 0.427 & 2.076 & 1.335 & 13.6+6.0 \\
Marigold-DC   & 4.924  & 2.595  & 0.505 & 0.151 & \textbf{1.465} & 0.434 & - \\
\textbf{Ours} & \textbf{4.387} & \textbf{2.177} & \textbf{0.366} & \textbf{0.138} & 1.471 & \textbf{0.422} & \textbf{0.9} \\
\bottomrule
\end{tabular}
\end{table}

\begin{table}[t]
\centering
\settablefont
\setlength{\tabcolsep}{6pt}
\caption{Additional ablation studies on the KITTI and VOID datasets.}
\label{tab:supp_ablation}
\begin{tabular}{lcccc}
\toprule
\multirow{2}{*}[-0.75ex]{Factor} &
\multicolumn{2}{c}{KITTI} &
\multicolumn{2}{c}{VOID} \\
\cmidrule(lr){2-3}\cmidrule(lr){4-5}
 & RMSE$\downarrow$ & MAE$\downarrow$ & RMSE$\downarrow$ & MAE$\downarrow$ \\
\midrule
Fitting: median               & 10.891 & 2.169 & 0.898 & 0.358 \\
Fitting: quantile matching    & 9.178  & 2.114 & 0.821 & 0.346 \\
Fitting: moment matching      & 8.756  & 1.983 & 0.815 & 0.329 \\
Fitting: least squares        & 7.013  & 1.802 & 0.791 & 0.307 \\
Domain: $z^{-1}$              & 6.782  & 1.794 & 0.614 & 0.238  \\
\midrule
Graph: global                 & 2.521  & 0.687 & 0.554 & 0.169      \\
Graph: graph-based (2D)       & 2.232  & 0.608 & 0.468 & 0.157      \\
Graph: graph-based (3D)       & 2.318  & 0.634 & 0.459 & 0.150      \\
With bilateral filtering      & \textbf{2.201}  & \textbf{0.597} & \textbf{0.452} & \textbf{0.148}      \\
\bottomrule
\end{tabular}
\end{table}

\begin{table}[t]
\centering
\settablefont
\setlength{\tabcolsep}{5.5pt}
\caption{Quantitative comparison with zero-shot stereo matching methods.}
\begin{tabular}{lcccc}
\toprule
\multirow{2}{*}{Method}
        & Middlebury & ETH3D & KITTI-12 & KITTI-15 \\
        & BP-2       & BP-1  & D1       & D1       \\
\midrule
CREStereo++~\cite{crestereo}        & 14.8 & 4.4 & 4.7 & 5.2 \\
DSMNet~\cite{dsmnet}             & 13.8 & 6.2 & 6.2 & 6.5 \\
Mask-CFNet~\cite{maskcfnet}         & 13.7 & 5.7 & 4.8 & 5.8 \\
HVT-RAFT~\cite{hvtraft}           & 10.4 & 3.0 & 3.7 & 5.2 \\
RAFT-Stereo~\cite{raft}        & 12.6 & 3.3 & 4.7 & 5.5 \\
Selective-IGEV~\cite{selective_igev}     &  9.2 & 5.7 & 4.5 & 5.6 \\
IGEV~\cite{igev}               &  8.8 & 4.0 & 5.2 & 5.7 \\
Former-RAFT-DAM~\cite{former_raft_dam}    &  8.1 & 3.3 & 3.9 & 5.1 \\
IGEV++~\cite{igev_pp}             &  7.8 & 4.1 & 5.1 & 5.9 \\
NMRF~\cite{nmrf}               &  7.5 & 3.8 & 4.2 & 5.1 \\
Ours               &  \textbf{6.1} & \textbf{2.8} & \textbf{3.6} & \textbf{5.0}\\
\bottomrule
\end{tabular}
\end{table}

\section{Experiments}
\label{Sect.additional_exp}
\subsection{Datasets}
For depth estimation, we use the KITTI~\cite{kitti}, NYU-Depth V2 (NYUv2)~\cite{nyu}, ScanNet~\cite{scannet}, ETH3D~\cite{eth3d}, and DIODE~\cite{diode} datasets. We follow the data splits adopted in the studies~\cite{marigold, depthmaster}. KITTI is a street-scene dataset with sparse metric depth captured by a LiDAR sensor, and we employ the Eigen test split~\cite{eigen_split}. The indoor dataset NYUv2 provides 654 images, while ScanNet contains 800 images. DIODE~\cite{diode} and ETH3D~\cite{eth3d} are high-resolution mixed indoor-outdoor datasets; in the splits, ETH3D contributes 454 images, and DIODE provides 325 indoor samples and 446 outdoor samples.

For depth completion, we use nuScenes~\cite{nuscenes}, DDAD~\cite{ddad}, Make3D~\cite{make3d}, DIODE~\cite{diode}, ETH3D~\cite{eth3d}, ScanNet~\cite{scannet}, VOID~\cite{void}, SUN-RGBD~\cite{sunrgbd}, IBims-1~\cite{ibims}, and HAMMER~\cite{hammer}. For nuScenes, we follow the preprocessing protocol and test split of prior work~\cite{Singh_2023_CVPR}. We discard samples with obvious projection errors. For DDAD, we follow the split and preprocessing of prior work~\cite{marigolddc} and use images at a resolution of 1216$\times$1936. For Make3D~\cite{make3d}, we adopt the split used in previous work~\cite{litemono}; due to the inherently low depth resolution of this dataset, the resulting MAE and RMSE are relatively large. For DIODE~\cite{diode}, ETH3D~\cite{eth3d}, and ScanNet~\cite{scannet}, we use the same splits as in our depth prediction experiments. For VOID~\cite{void}, we again follow the split proposed in prior work~\cite{piccinelli2024unidepth}. For SUN-RGBD~\cite{sunrgbd}, we use the split that excludes images overlapping with NYUv2~\cite{nyu}, resulting in approximately 4.4k images. For HAMMER~\cite{hammer}, we randomly sample 400 images for testing. For IBims-1~\cite{ibims}, we use the official evaluation split with 100 images. nuScenes~\cite{nuscenes}, DDAD~\cite{ddad}, and VOID~\cite{void} provide sparse depth maps directly. For the remaining datasets, we simulate sparse depth by randomly retaining 0.01\%–0.1\% depth points. We apply noisy random sampling, where points are uniformly sampled at varying densities and 10\%–20\% of them are perturbed with noise. For datasets acquired with line-scan LiDAR sensors, we use LiDAR-simulated sampling.

\begin{figure*}[!t]
	\centering
	\includegraphics[width=0.99\textwidth]{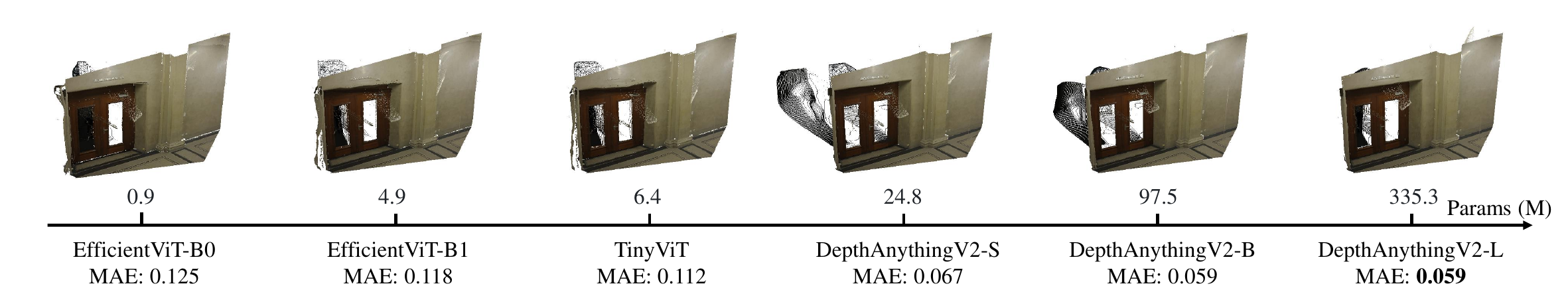}
	\caption{Qualitative results across different depth foundation models on the ETH3D Indoor dataset.
    }
\label{fig.foundation}
\end{figure*}

\begin{figure*}[!t]
	\centering
	\includegraphics[width=0.99\textwidth]{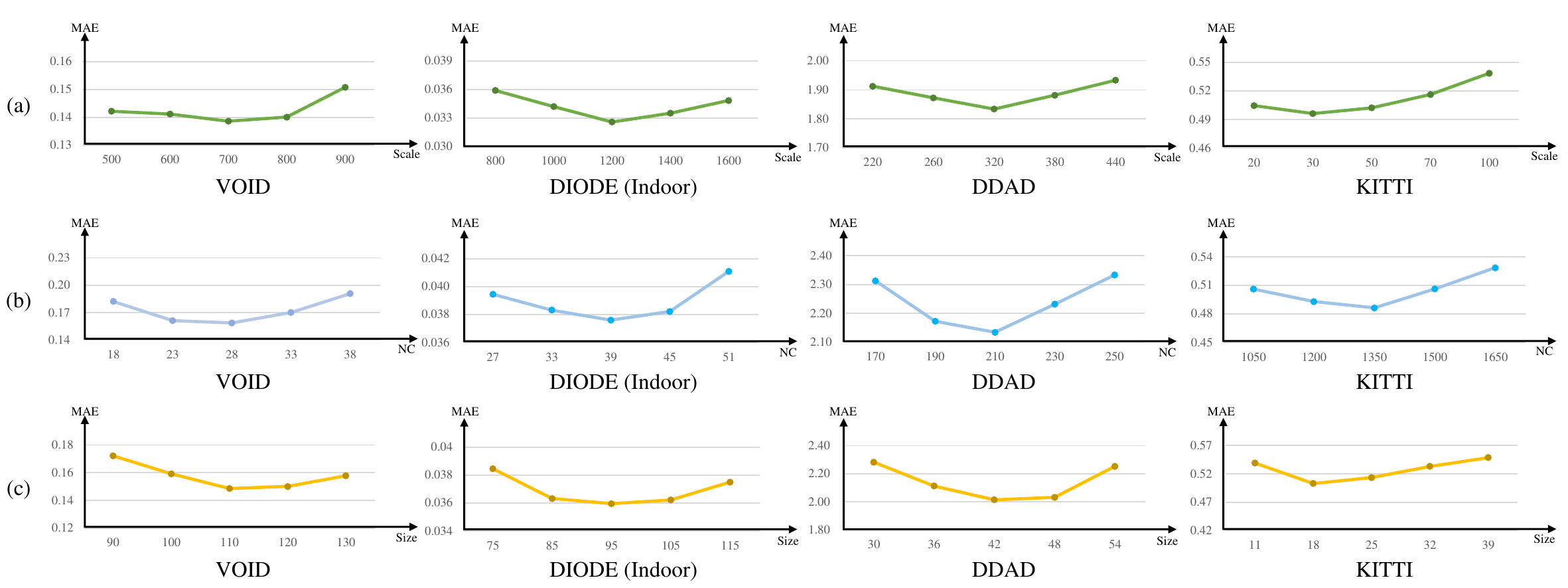}
	\caption{MAE comparison under different superpixel segmentation algorithms. The subfigures correspond to (a) the Felzenszwalb algorithm~\cite{felzen}, (b) the SLIC algorithm~\cite{slic}, and (c) the LSC algorithm~\cite{LSC}. In (b), ``NC'' denotes the number of components in SLIC~\cite{slic}. Under a fixed number of 3D seed points, we search over the hyperparameters of each algorithm to minimize MAE. VOID and DIODE are indoor datasets, while DDAD and KITTI are outdoor datasets.}
\label{fig.superpixel}
\end{figure*}

\subsection{The Details of Comparison with SoTA Methods}
In our main paper, for the depth estimation baselines, the quantitative results reported in our tables (the section without our method) are taken from the original studies~\cite{marigold, depthmaster}. In our main paper, for depth completion baselines, we follow the protocol of prior work~\cite{dmd3c, ognidc} and conduct zero-shot evaluation for CFormer~\cite{cformer}, BP-Net~\cite{bpnet}, and LRRU~\cite{lrru} by using weights trained on KITTI for outdoor datasets and weights trained on NYUv2 for indoor datasets. For DMD$^3$C~\cite{dmd3c}, we use the official weights trained on a large-scale dataset for zero-shot evaluation. It is worth noting that the pretraining of DMD$^3$C~\cite{dmd3c} includes nuScenes, which partly explains its superior performance on the nuScenes benchmark. Although DMD$^3$C~\cite{dmd3c} and BP-Net~\cite{bpnet} share similar network architectures, we observe that this large-scale pretraining substantially improves the generalization ability of DMD$^3$C. For PromptDA~\cite{promptda}, since the official weights are trained on a specific dataset and exhibit poor generalization, we reimplement the method. We configure the denoising process of Marigold-DC~\cite{marigolddc} within a reasonable range.

\begin{figure}[!t]
	\centering
	\includegraphics[width=0.49\textwidth]{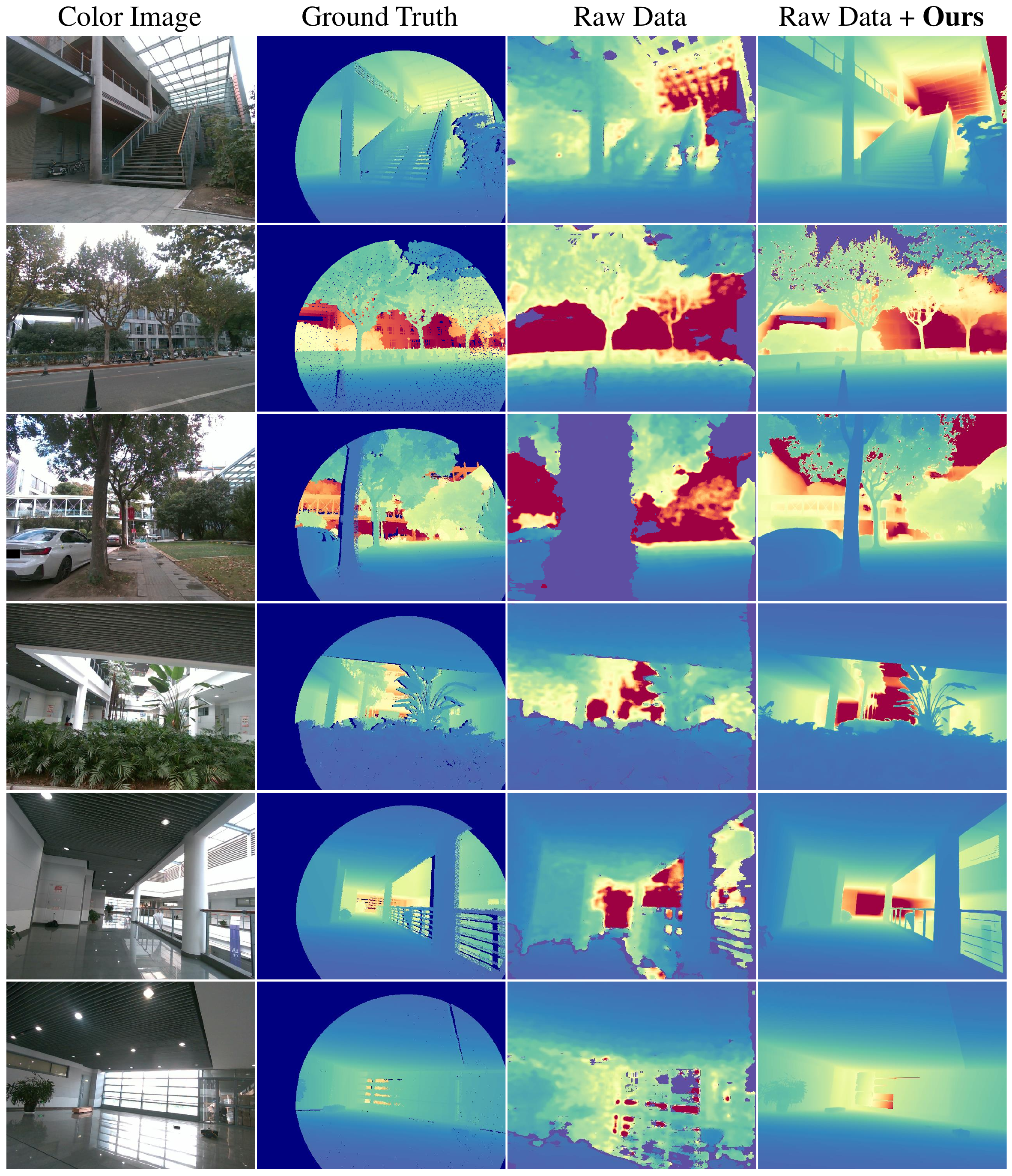}
	\caption{Qualitative results for the rectification of commonly used range cameras.}
\label{fig.big}
\end{figure}

Furthermore, Table~\ref{tab:supp_dc} provides additional comparisons on multiple depth completion datasets, complementing comparisons for the main paper. In Table~\ref{tab:supp_mde}, we fine-tune UniDepth~\cite{piccinelli2024unidepth} on the downstream datasets to obtain 3D seeds. This setup allows our method to use only monocular color images as input.
Fig.~\ref{fig.qual} shows qualitative comparisons among our method and SoTA depth completion methods on multiple datasets. Fig.~\ref{fig.dc_illu} illustrates the process of reducing local scale inconsistencies. As shown in the figure, the original predictions of DepthAnythingV2~\cite{depthanythingv2} exhibit noticeable local inconsistencies around the tree trunks, while our segment-wise recovery strategy reduces these inconsistencies. This qualitative evidence further validates the motivation underlying our method. In addition, Fig.~\ref{fig.supp_rebuttal} shows that when the color image and the LiDAR projected sparse depth are misaligned, a condition commonly encountered in outdoor datasets, our method can still robustly recover a well-aligned dense depth map.

We also provide a quantitative comparison with SPN-based methods in Table~\ref{tab:spn}. Unlike SPN-based methods, we derive a discontinuity-aware propagation metric from second-order residuals (Proposition 1), leading to a geodesic shortest path formulation that can be solved efficiently via dynamic programming. Beyond not requiring training, our method has three advantages. (1) Better interpretability: $d_\phi$ has a clear geometric meaning as a geodesic cost that penalizes discontinuity crossings, unlike implicitly learned affinities. (2) Better cross-domain generalization: our geometry-driven cost avoids affinity learning and is therefore less sensitive to domain shifts. (3) Higher efficiency and lower computational overhead: our approach is lightweight and plug-and-play, requiring no additional neural network to produce pixel-wise propagation.

\subsection{Ablation Study}

\begin{figure}[!t]
	\centering
	\includegraphics[width=0.49\textwidth]{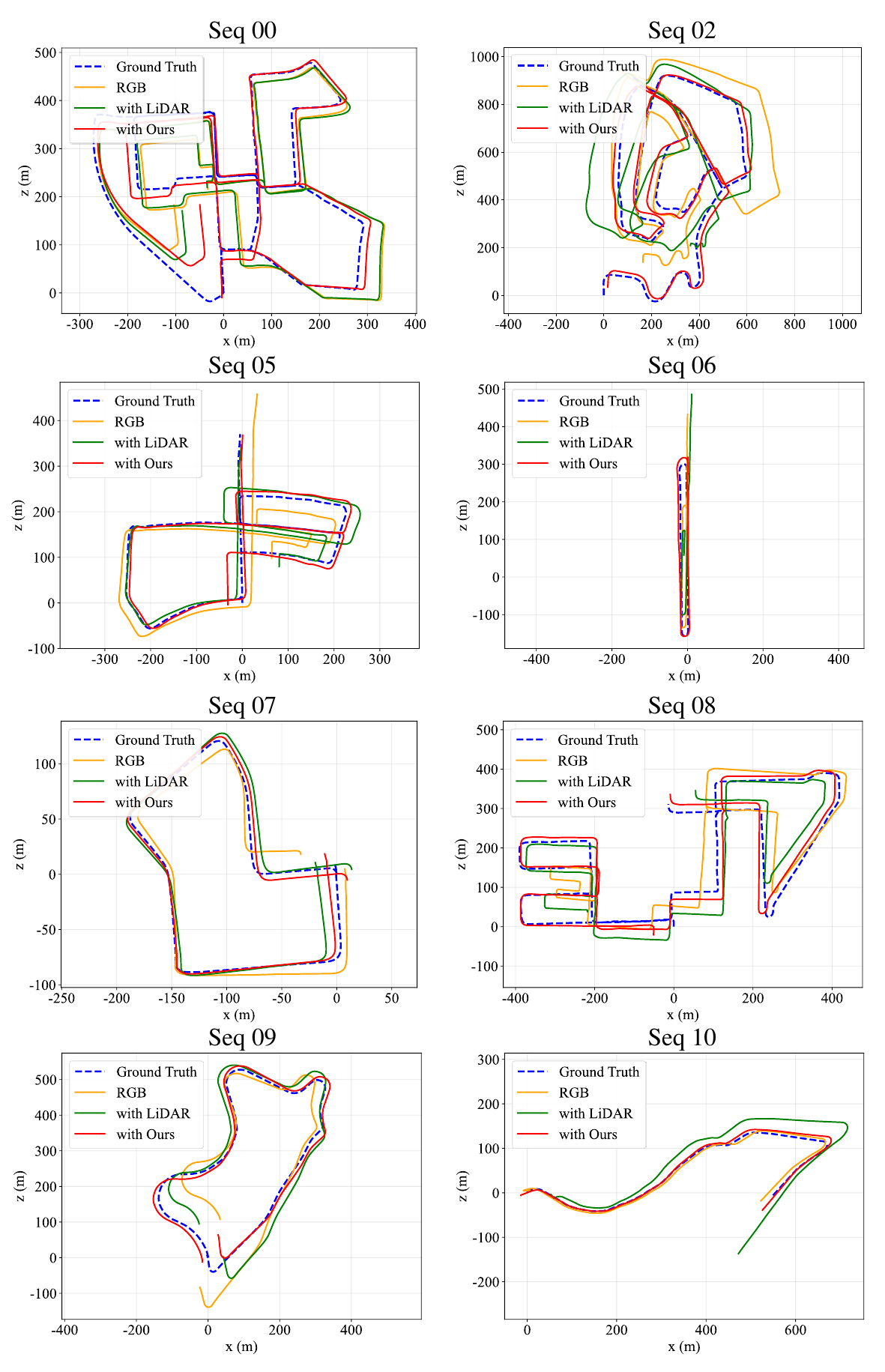}
	\caption{Additional qualitative results on the KITTI Odometry benchmark.}
\label{fig.supp_kitti_odom}
\end{figure}

In this section, we first describe the ablation setting used in the main paper. In Fig. 4 of the main paper, the ETH3D and DIODE results are computed only on their indoor subsets. The purpose of this ablation is to analyze how the number of 3D seed points (NP) influences performance under different scene types (indoor versus outdoor). In contrast, Table 1 of the main paper reports results on the full splits, where both indoor and outdoor scenes of ETH3D and DIODE are included. Consequently, the results are different between these two settings.

The main paper shows the effectiveness of our segment-wise recovery and pixel-wise refinement. Since the supplementary material describes the detailed methods for computing the per-segment calibration function and explains the iterative bilateral filtering used in segment propagation and graph optimization, we present more detailed ablation studies for segment-wise recovery. In Table~\ref{tab:supp_ablation}, the upper part of the table compares the results under different calibration functions. The lower part of the table compares whether the distance term $\|\boldsymbol{c}_i - \boldsymbol{c}_j\|$ in the sparse graph optimization uses 2D or 3D coordinates. We also demonstrate the benefit of adding our bilateral filtering.

 Fig.~\ref{fig.foundation} compares the results of our method across depth foundation models with different numbers of parameters for the same input. The figure shows point cloud visualizations on the ETH3D Indoor dataset. EfficientViT-B0, EfficientViT-B1, and TinyViT are models that we obtain by distillation. These lightweight depth foundation models achieve MAE within an acceptable range and provide a good balance between accuracy and efficiency, which makes them suitable for other downstream tasks. The MAE difference between DepthAnythingV2-S and DepthAnythingV2-B is relatively minor, and the difference between DepthAnythingV2-B and DepthAnythingV2-L is negligible. This result also indicates that our method does not depend on high-capacity relative depth estimators to achieve accurate metric depth estimation.

Moreover, in Fig.~\ref{fig.superpixel}, we search for the optimal hyperparameters of the superpixel segmentation algorithms to minimize MAE, and we also compare different superpixel algorithms under a fixed number of 3D seed points. The optimal parameters of the three superpixel methods are related. For example, for (a) the Felzenszwalb algorithm~\cite{felzen}, if the optimal scale produces $N$ distinct segments, then for (b) the SLIC algorithm~\cite{slic}, the optimal number of components is also close to $N$. If $A = H \times W / N$ denotes the approximate area of each segment, then for (c) the LSC algorithm~\cite{LSC}, the optimal value of the size parameter lies near $\sqrt{A}$. In addition, the MAE does not change significantly across different hyperparameter settings, and the optimal MAE values of the three algorithms are relatively close. These observations demonstrate the robustness of our method. Therefore, in the ablation studies in the main paper, we use the Felzenszwalb algorithm~\cite{felzen}, as it achieves the best overall performance on both indoor and outdoor datasets.

\begin{figure}[!t]
	\centering
	\includegraphics[width=0.49\textwidth]{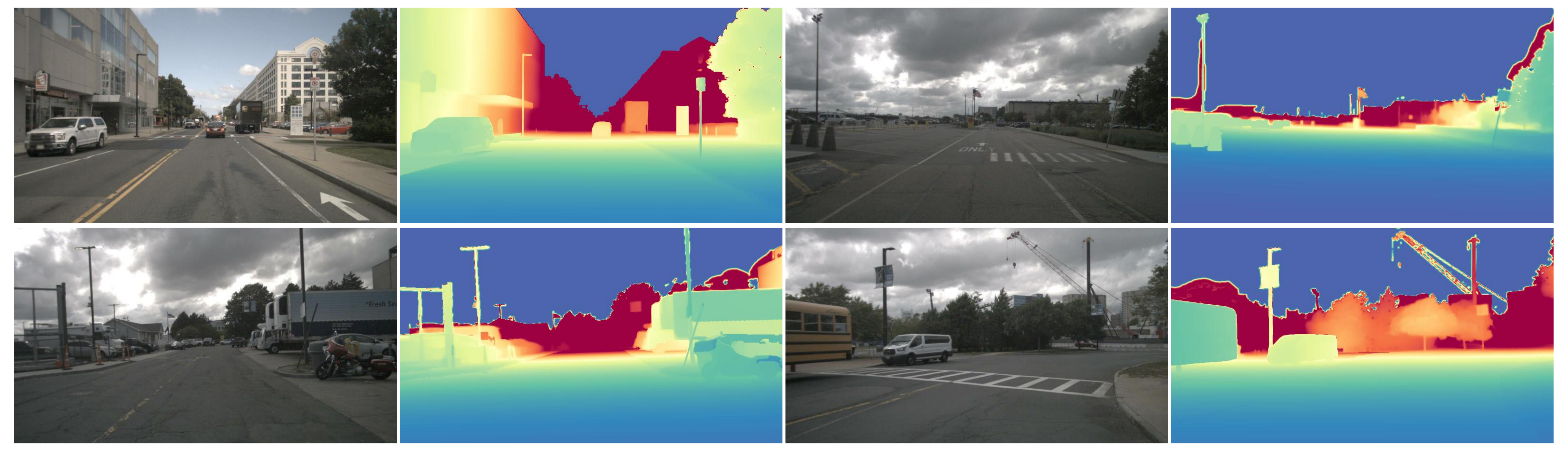}
	\caption{Qualitative results on our method for the Radar depth completion task.}
\label{fig.radar}
\end{figure}

\subsection{Applications}

Fig.~\ref{fig.big} presents qualitative results obtained by applying our method to depth rectification for commonly used range cameras. Our method substantially improves the quality of raw depth data by filling missing regions, sharpening depth textures, and preserving object boundaries. Comparing our results with the ground-truth visualizations, we observe that many objects that appear blurry in the raw data become clearer and exhibit more accurate depth values. Fig.~\ref{fig.supp_kitti_odom} shows additional SLAM results from the KITTI Odometry benchmark. Compared with using DROID-SLAM~\cite{droid} directly on raw point clouds, incorporating our method clearly improves the results.

In the main paper, we show that our method can be integrated with vision foundation models and applied to multi-view stereo. Table~\ref{fig.supp_kitti_odom} further demonstrates that our method can also be applied to stereo matching, where it achieves competitive performance. We use the high-confidence matches from~\cite{loftr} as disparity inputs and reconstruct dense disparity maps. Fig.~\ref{fig.radar} presents qualitative radar depth completion results on the nuScenes dataset. These experiments demonstrate that our method generalizes to different input modalities, thereby confirming its effectiveness on downstream tasks.

\small
\normalem
\bibliographystyle{ieeenat_fullname}
\bibliography{ref}

\end{document}